%% file: root.tex
\documentclass[letterpaper, 10 pt, journal, twoside]{IEEEtran}  

\IEEEoverridecommandlockouts                              

\usepackage{graphics} 
\usepackage{epsfig} 
\usepackage{mathptmx} 
\usepackage{amsmath} 
\usepackage{amssymb}  
\usepackage{cite}

\usepackage[normalem]{ulem}
\usepackage{picinpar}
\usepackage{url}
\usepackage{flushend}
\usepackage{colortbl}
\usepackage{soul}
\usepackage{multirow}
\usepackage{pifont}
\usepackage{color}
\usepackage{alltt}
\usepackage{hyperref}
\usepackage{siunitx}
\usepackage{breakurl}
\usepackage{epstopdf}
\usepackage{pbox}
\usepackage{subfigure}
\usepackage{threeparttable}
\usepackage{graphicx}
\usepackage{bm} 
\usepackage{adjustbox}
\usepackage{booktabs}
\usepackage[utf8]{inputenc}
\usepackage{booktabs}
\usepackage{threeparttable}
\usepackage{textcomp}
\usepackage{hyperref}
\usepackage{booktabs}
\usepackage{float}
\usepackage{multirow}
\usepackage{multicol}
\usepackage[table,xcdraw]{xcolor}
\usepackage{colortbl}
\usepackage{makecell}
\usepackage{enumitem}
\usepackage{xcolor} 

\newcommand{\pmv}[2]{\ensuremath{#1{\pm}#2}}

\title{HyVGGT-VO: Hybrid Dense Visual Odometry With Classical VO and Feed-Forward Model}

\author{
Junxiang Pang,
Lipu Zhou,
Baojie Chen,
and Zhenzhong Wei%
\thanks{Manuscript received: April 1, 2026; Revised: August 14, 2026; Accepted: July 30, 2026. This paper was recommended for publication by Editor Ayoung Kim upon evaluation of the Associate Editor and Reviewers’ comments. This work was supported in part by the Fundamental Research Funds for the Central Universities under Grant
501GWJH2024117003 and in part by the National Science Foundation
of China under Grant 52127809. \textit{(Corresponding author: Lipu Zhou.)}}%
\thanks{The authors are with the Key Laboratory of Precision
Opto-Mechatronics Technology, Ministry of Education, Beijing 100191, China, and also with the School of Instrumentation and Optoelectronic Engineering, Beihang University, Beijing 100191, China (e-mail: pangjunxiang@buaa.edu.cn; zhoulipu@buaa.edu.cn;
chenbaojie@buaa.edu.cn; zhenzhongwei@buaa.edu.cn).}%
\thanks{Digital Object Identifier (DOI): 10.1109/LRA.2026.3728304}
}

\begin{document}

\bstctlcite{IEEEexample:BSTcontrol}

\maketitle

\begin{abstract}
\input{article/abstract}
\end{abstract}


\section{Introduction}
\input{article/introduction}

\section{Related Work}
\input{article/relatedworks}

\section{Methodology}
\input{article/methodology}

\section{Experiments}
\input{article/experiments}

\section{Conclusion}
\input{article/conclusion}



\bibliographystyle{IEEEtran}
\bibliography{ref}
\end{document}

%% file: article/abstract.tex
Dense visual odometry (VO), which provides pose estimation and dense 3D reconstruction, serves as the cornerstone for applications ranging from robotics to augmented reality. Recently, feed-forward models have demonstrated remarkable capabilities in dense mapping. However, when these models are used in dense visual SLAM systems, their keyframe-based processing often leads to sparse pose outputs, limiting continuous per-frame pose estimation. In contrast, traditional sparse methods provide high-frequency pose outputs but lack the capability for dense reconstruction. To address these limitations, we propose HyVGGT-VO, a novel framework that combines the computational efficiency of sparse VO with the dense reconstruction capabilities of feed-forward model. To the best of our knowledge, this is the first work to tightly couple a traditional VO frontend with VGGT. Specifically, we design an adaptive hybrid tracking frontend that dynamically switches between traditional optical flow and the VGGT model to ensure robustness. Furthermore, we introduce a hierarchical optimization framework that jointly refines VO poses and the scale of VGGT predictions to ensure global scale consistency. Under the odometry setting, compared with VGGT-SLAM 2.0, HyVGGT-VO reduces the average ATE by approximately 91\% on EuRoC MAV and 20\% on TUM RGB-D, and decreases the average relative translation and rotation errors by 46\% and 92\% on KITTI Seq. 00–05, respectively, while achieving 16.1 Hz pose output and 3.2 Hz dense-mapping updates with 6.5 GB of memory on an 8 GB laptop GPU. Our code is available at \url{https://geneta2580.github.io/HyVGGT-VO.io}.

%% file: article/introduction.tex
\IEEEPARstart{D}{ense} visual odometry (VO), which simultaneously provides pose estimation and dense scene representation, serves as the cornerstone for modern spatial perception. This dense representation is crucial for various downstream tasks: it provides rich texture and 3D structural information for Vision-Language Models (VLMs), while providing essential localization and mapping data for motion planning in robotics.

\begin{figure}[t]
\centering
\includegraphics[width=1.0\linewidth]{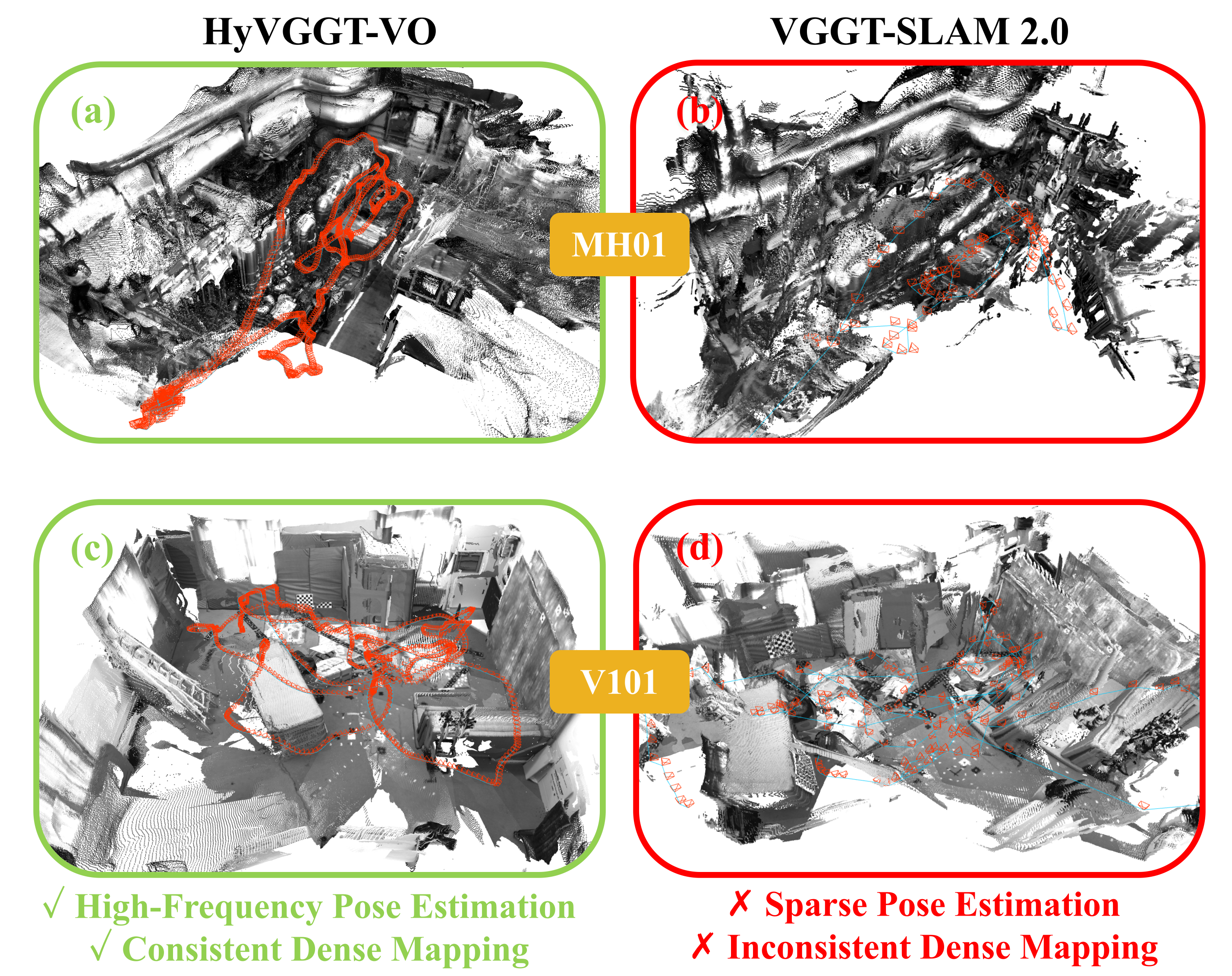}
\vspace{-0.7cm}
\caption{Motivation and qualitative comparison of dense 3D reconstructions on the EuRoC dataset. VGGT-SLAM 2.0~\cite{vggt-slam2.0} yields sparse and delayed keyframe pose outputs, and exhibits severe reconstruction inconsistencies in the MH01 (b) and V101 (d) sequences. In contrast, our method produces globally consistent dense maps while delivering high-frequency poses and smoother trajectories as shown in MH01 (a) and V101 (c).}
\label{fig:motivation}
\vspace{-0.5cm}
\end{figure}

Traditional sparse visual simultaneous localization and mapping (SLAM) methods provide high computational efficiency and high-frequency pose outputs, but lack dense scene representation. Meanwhile, existing traditional dense visual SLAM methods \cite{dtam, dso} struggle with robustness, failing in environments with dynamic illumination or regions lacking texture. To overcome these fragility issues, deep learning approaches, such as DeepV2D \cite{deepv2d} and DROID-SLAM \cite{droid-slam}, reformulate the tracking and optimization pipeline using deep iterative architectures to enhance robustness. However, their core optimization heavily relies on alternating updates or dense bundle adjustment (DBA) layers, which consume excessive computational resources.

To simplify the 3D reconstruction pipeline, feed-forward models have emerged \cite{dust3r, mast3r, lrm, mvsformer++, monst3r, meshformer, map-anything, pi3, ttt3r, depth-anything3, vggt-omega, zipmap, amb3r, pow3r}. These models can directly regress camera poses and dense geometry from images via a single forward inference. This significantly reduces system complexity and has become a promising new paradigm \cite{feedforwad-review}. As a significant advancement in this domain, the recent VGGT model \cite{vggt} employs an architecture that directly predicts comprehensive 3D attributes, delivering high accuracy for camera poses and 3D geometry at high inference speeds.

Given these advantages, VGGT has recently been integrated into dense visual SLAM systems. For instance, methods such as VGGT-SLAM \cite{vggt-slam2.0}, VGGT-Long \cite{vggt-long}, and VGGT-Motion \cite{vggt-motion} alleviate high GPU memory usage by partitioning long trajectories into sub-graphs and aligning point clouds from overlapping frames. Although this strategy reduces memory usage on long sequences, its reliance on keyframe selection yields sparse pose outputs and delayed pose output. In addition, inaccuracies in feed-forward pose or geometry predictions may affect the consistency between independently processed sub-graphs, as illustrated in Fig.~\ref{fig:motivation}.

Taken together, there is a strong complementarity between feed-forward models and traditional sparse VO. While the former excels at dense reconstruction despite computational latency, the latter excels at high-frequency pose estimation and computational efficiency. Motivated by this, we propose HyVGGT-VO, a hybrid framework designed to exploit the strengths of both paradigms via asynchronous fusion.

Our main contributions can be summarized as follows:
\begin{itemize}[leftmargin=*]
\item \textbf{A Novel Hybrid VO Architecture:} We propose HyVGGT-VO, a tightly coupled framework that combines the high-frequency pose estimation of classical VO with the dense mapping capability of the VGGT model, balancing pose rate, dense reconstruction, and GPU memory efficiency.

\item \textbf{Adaptive Uncertainty-Aware Frontend:} We design an adaptive hybrid frontend in which classical VO provides the primary pose estimates, while VGGT is invoked only under challenging conditions when classical tracking fails, incorporating an edge-aware uncertainty model to adjust measurement noise based on network confidence.

\item \textbf{Trajectory-Based Scale Alignment and Hierarchical Backend:} To improve the scale consistency of feed-forward models, we introduce a trajectory-based $Sim(3)$ alignment strategy. Furthermore, we design a hierarchical optimization backend that integrates local bundle adjustment (BA) and local pose graph optimization (PGO), where the scale is explicitly parameterized as an optimization variable.

\end{itemize}

%% file: article/relatedworks.tex
\subsection{Traditional Visual SLAM} 
Traditional visual SLAM frameworks are generally categorized into indirect, semi-direct, and direct approaches. Prominent indirect feature-based methods include ORB-SLAM \cite{orb-slam, orb-slam2, orb-slam3} and OV$^2$SLAM \cite{ov2-slam}. Semi-direct approaches are exemplified by SVO \cite{svo}, whereas direct methods, operating on raw pixel intensities, are represented by DSO \cite{dso}. While providing high-frequency pose estimation, traditional methods suffer from degraded robustness in challenging scenarios, such as textureless regions or severe illumination changes.

\subsection{Learning-based Visual SLAM} 
To address the robustness limitations of traditional visual SLAM, learning-based approaches have been introduced. Early works like DeepFactors \cite{deepfactors} integrate learned compact depth codes into a probabilistic factor graph, while DeepV2D \cite{deepv2d} employs differentiable motion stereo blocks to alternate between depth estimation and pose refinement. Differing from optimization-based schemes, TartanVO \cite{tartan-vo} demonstrates the generalization potential of pure end-to-end learning by directly regressing odometry from optical flow trained on massive synthetic datasets. More recently, DROID-SLAM \cite{droid-slam} has achieved notable robustness through a recurrent iterative update module based on dense optical flow. However, DROID-SLAM relies on a complex recurrent network architecture. This substantial computational demand limits its applicability to resource-constrained platforms.

\subsection{Feed-forward Dense Mapping}
In the realm of 3D dense reconstruction, feed-forward models have emerged as a promising paradigm to simplify the intricate pipelines of traditional algorithms. These models can directly regress 3D geometric parameters and camera poses via a single forward inference. DUSt3R \cite{dust3r} is a pioneering work in this domain, reformulating multi-view stereo as a point map regression problem to enable reconstruction without prior camera calibration. Based on this, MASt3R \cite{mast3r} incorporates a matching-based strategy to improve the alignment of local features. Meanwhile, Spann3R \cite{spann3r} and Fast3R \cite{fast3r} extend these concepts to sequential image processing. Subsequently, VGGT \cite{vggt} has presented an architecture that directly predicts comprehensive 3D attributes. It treats multi-view localization and dense reconstruction as a joint prediction problem, regressing camera intrinsics, extrinsics, depth maps, and dense point clouds from an input image set. 

To leverage the robust dense reconstruction capabilities of feed-forward models, recent works have integrated them into visual SLAM frameworks\cite{mast3r-slam, vista-slam}. For example, MASt3R-SLAM extends the MASt3R to handle long sequences. Following this trend, subsequent works have integrated the advanced VGGT model into SLAM systems. These VGGT-based methods typically partition continuous sequences into sub-graphs. For instance, VGGT-Long \cite{vggt-long} and VGGT-Motion \cite{vggt-motion} perform local registration by estimating $Sim(3)$ transformations between overlapping point clouds. To further address structural distortions caused by feed-forward predictions, VGGT-SLAM \cite{vggt-slam} adopts a flexible $SL(4)$ alignment scheme, while its extension \cite{vggt-slam2.0} refines keyframe poses within each sub-graph to improve local accuracy.

While these sub-graphs merging strategies can handle long sequences, maintaining consistent scale across independently processed sub-graphs remains challenging when the feed-forward predictions contain local pose or geometric errors. Furthermore, due to the computational cost of VGGT, these methods typically process sparse keyframes, which limits their ability to provide per-frame pose outputs. To address these limitations, our work integrates the computational efficiency and local covisibility constraints of traditional VO into the VGGT framework, mitigating scale ambiguity while maintaining high-frequency pose estimation.

%% file: article/methodology.tex
\subsection{System Overview} 
\begin{figure*}[!t]
  \centering
  \includegraphics[width=0.95\linewidth]{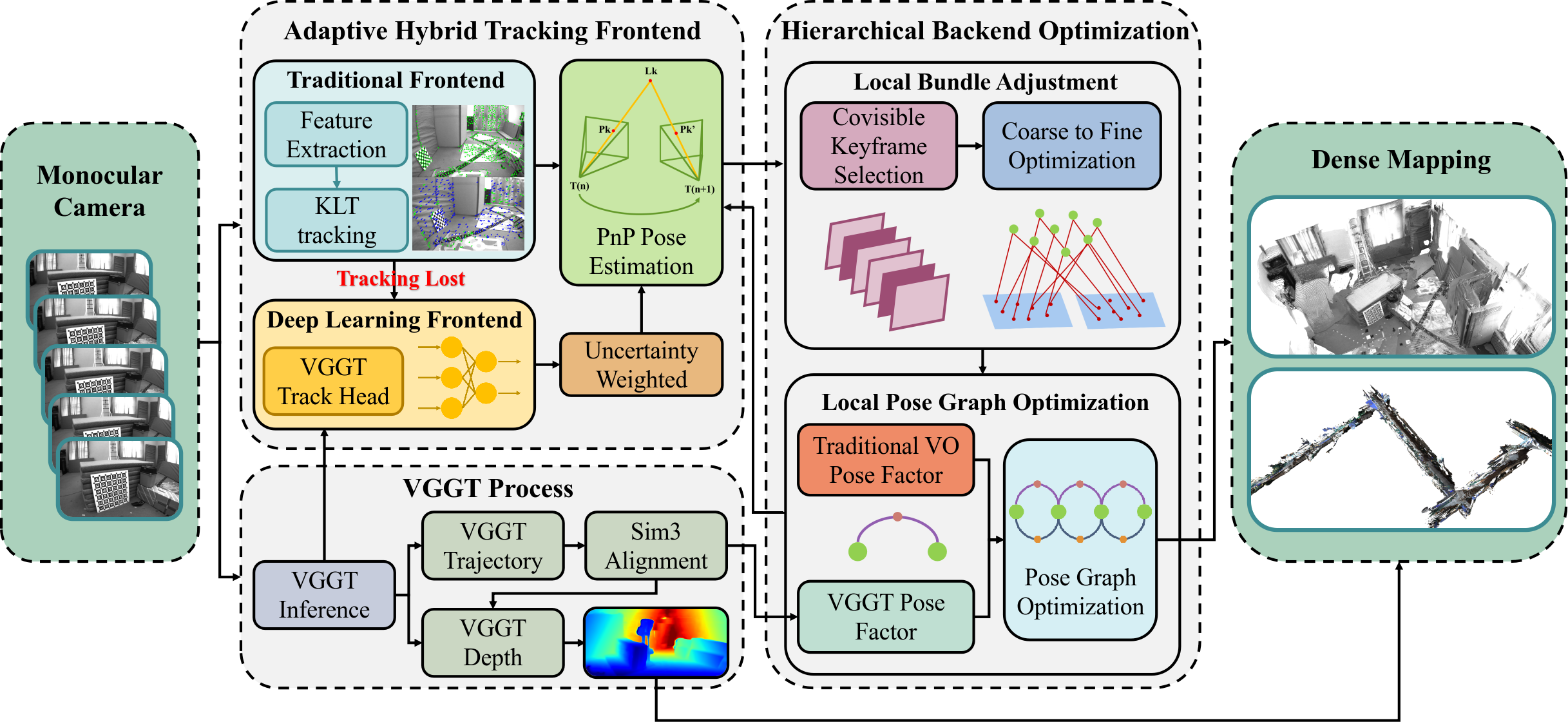}
  \vspace{-0.1cm}
  \caption{Overall architecture of HyVGGT-VO. The system couples a KLT-based sparse VO frontend with VGGT tracking and an asynchronous hierarchical backend for local BA and VGGT-prior-based local PGO.}
  \label{fig:system}
  \vspace{-0.5cm}
\end{figure*}

The overall architecture of HyVGGT-VO is illustrated in Fig.~\ref{fig:system}. The system consists of a hybrid tracking frontend and a hierarchical backend optimization module.

\subsection{Adaptive Hybrid Tracking Frontend}
\textbf{Primary Tracking with Priors:}
HyVGGT-VO uses a pyramidal KLT-based tracker as the primary frontend for efficient pose estimation, following OV$^2$SLAM~\cite{ov2-slam}. We maintain both triangulated 3D map points and untriangulated 2D feature tracks. For triangulated points, a constant-velocity motion model projects the previous map points into the current frame, providing search priors for KLT tracking. The resulting correspondences are used for pose estimation, while untriangulated tracks are retained for later triangulation.

\textbf{VGGT Robust Tracking:}
Although classical VO is computationally efficient, it becomes unreliable under challenging conditions. HyVGGT-VO therefore invokes VGGT when classical tracking degrades. We compute the KLT tracking ratio as $\rho_t=N^{\mathrm{succ}}_t/N^{\mathrm{prev}}_t$, where $N^{\mathrm{prev}}_t$ and $N^{\mathrm{succ}}_t$ denote the valid 3D points before and after KLT tracking, respectively. If $\rho_t<\tau$ ($\tau=0.8$), VGGT-based tracking is triggered and the recovered correspondences are used for PnP. If PnP still has insufficient inliers, we use optical flow to select keyframes and group the selected frames into fixed-size sub-graphs of 10 frames for VGGT inference. The relative scale of the newly inferred sub-graph is estimated from the median ratio between valid depth predictions in the overlapping frame, and the scaled pose and depth predictions are then propagated from the overlapping keyframe as the anchor. After each recovery, the next frame again prioritizes KLT tracking and returns to normal tracking once sufficient correspondences are obtained.

\textbf{Outlier Rejection and Uncertainty Integration:}
Since the feed-forward predictions of the VGGT model do not strictly satisfy geometric constraints, rigorous verification is crucial. We first enforce geometric consistency through a RANSAC-based epipolar check, followed by P3P-RANSAC and PnP optimization to reject outliers.

Furthermore, we introduce a dynamic noise adjustment strategy that integrates the confidence score $u\in(0,1]$ predicted by the VGGT model with an edge penalty mechanism. VGGT continues tracking features that have already moved outside the camera's field of view. Such erroneous predictions tend to accumulate near image boundaries and reduce tracking quality. We therefore down-weight boundary observations by adaptively scaling their measurement noise in both frontend pose estimation and backend optimization. For feature $k$, the noise scale is defined as $\sigma_k = u_k^{-1}\sigma_b\eta_e(\mathbf{p}_k)$, where $\sigma_b$ is the baseline measurement noise, $u_k$ denotes the feature uncertainty term, and $\eta_e(\mathbf{p}_k)$ is a spatial penalty that increases the noise of features near image boundaries. It applies a substantial penalty factor $k_p$ to points within a margin $\delta$ ($\delta$ =10 pixels) of the image boundary:
\begin{equation}
\eta_{e}(\mathbf{p}_k) = 
\begin{cases} 
k_p & \text{if } \text{dist}(\mathbf{p}_k, \textit{\textbf{B}}) < \delta \\
1 & \text{otherwise}
\end{cases},
\end{equation}
where $\text{dist}(\cdot, \textit{\textbf{B}})$ represents the pixel distance to the image boundary. In our implementation, we employ a large penalty factor $k_p$ ($k_p = 10$) to effectively down-weight edge observations. This adaptive noise model is utilized to construct the covariance matrix $\bm{\Sigma}_k = \text{diag}(\sigma_k^2, \sigma_k^2)$ for the cost functions in both pose estimation and local BA optimization.

\textbf{Pose Estimation:}
We employ a constant velocity model to predict the initial pose $\mathbf{T}_{wc}$. Subsequently, a PnP problem is formulated using the verified 2D-3D correspondences to optimize the current camera pose. This optimization minimizes the weighted reprojection error by incorporating the adaptive covariance $\sigma_{k}$ defined in (1) and employing a Huber robust kernel to mitigate the impact of outliers.

\subsection{Scale Alignment Strategy}
Existing VGGT-based methods~\cite{vggt-long,vggt-slam} partition sequences into overlapping sub-graphs to mitigate memory overhead, and often use overlapping point clouds or local geometric predictions to align independently processed segments. Such alignment can be affected by local geometric distortions or inaccurate feed-forward pose estimates, which may reduce scale consistency across sub-graphs.

To compute the scale factor, we employ a trajectory alignment method. We align the trajectory segment predicted by VGGT to the corresponding traditional VO trajectory, which generally provides a stable reference for the scale of the network outputs.

Specifically, let $K_i$ denote the set of keyframes in the $i$-th VGGT inference sub-graph. We extract the corresponding poses computed by the traditional VO, denoted as $\mathbf{T}_{o} = \{ \mathbf{T}_{o}^k \mid k \in K_i \}$, and the raw poses predicted by VGGT, denoted as $\mathbf{T}_{v} = \{ \mathbf{T}_{v}^k \mid k \in K_i \}$. 
It is worth noting that the raw translations predicted by VGGT are typically scale-normalized and exhibit small magnitudes, which can lead to numerical instability during optimization. To mitigate this, we introduce a pre-scaling factor $\gamma$ ($\gamma$ = 20) to normalize the magnitude of the network predictions before alignment.

We formulate the alignment problem as finding the optimal similarity transformation $\mathbf{S} \in Sim(3)$ that aligns the pre-scaled network trajectory to the VO trajectory. This is solved efficiently using the Umeyama algorithm \cite{umeyama} by minimizing the least-squares error of the translation components:

\begin{equation}
\min_{s, \mathbf{R}, \mathbf{t}} \sum_{k \in K_i} \left\| \mathbf{t}_{vo}^k - \left( s \cdot \mathbf{R} \cdot (\gamma \cdot \mathbf{t}_{vggt}^k) + \mathbf{t} \right) \right\|^2,
\end{equation}
where $\mathbf{t}_{vo}^k$ and $\mathbf{t}_{vggt}^k$ are the translation vectors of the corresponding poses. The optimization yields the relative scale factor $s$, the rotation $\mathbf{R}$, and the translation $\mathbf{t}$. In our hybrid framework, we primarily utilize the computed scale $s_{f} = s \cdot \gamma$ to correct the depth and translation of the current VGGT sub-graph, ensuring it acts as a consistent geometric prior for the subsequent factor graph optimization.

\subsection{Hierarchical Backend Optimization}

\begin{figure}[!t]
  \centering
  \includegraphics[width=0.96\linewidth]{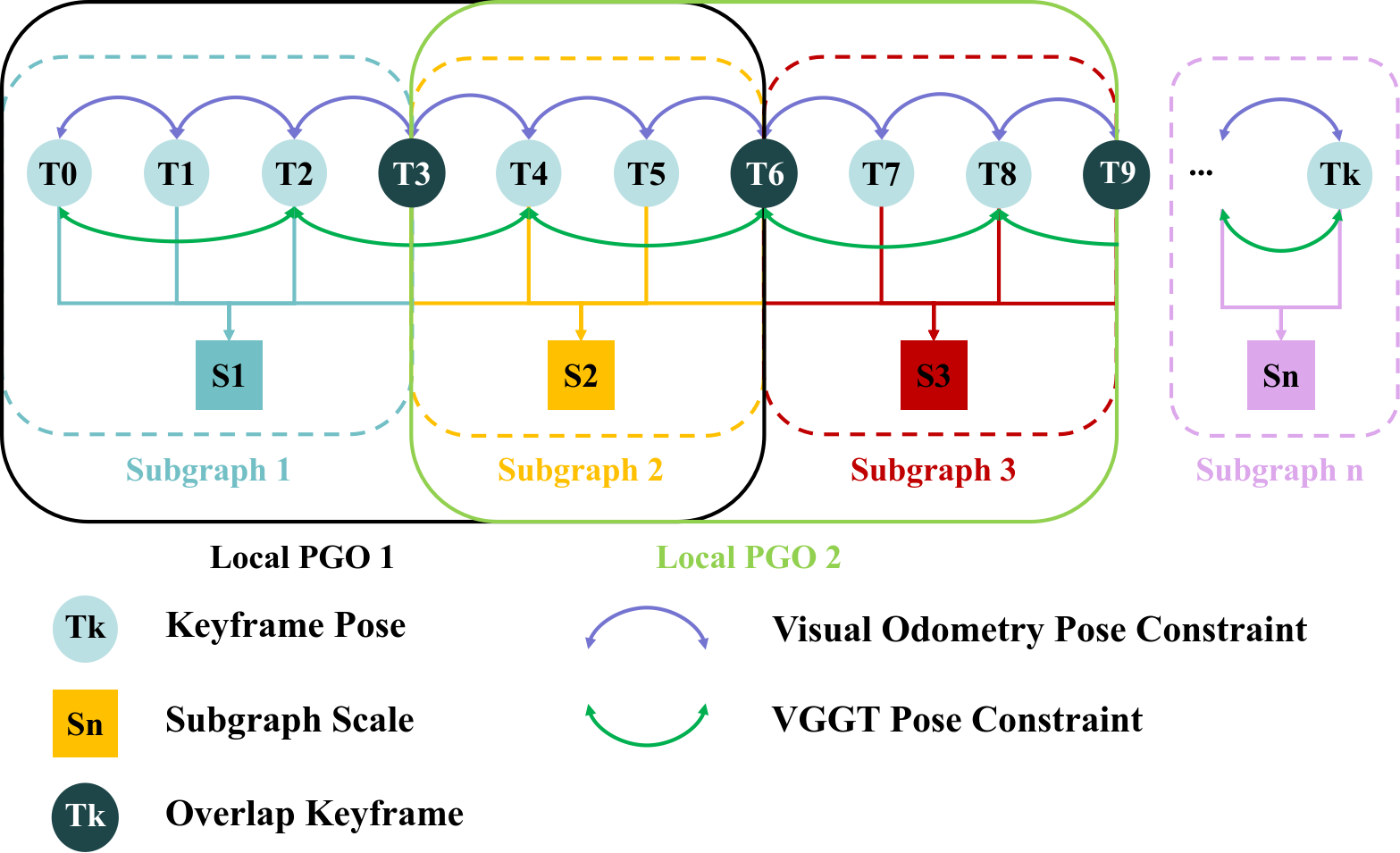}
   \vspace{-10pt}
  \caption{Structure of our local PGO. The first frame of the first sub-graph is fixed as the world reference frame. The sub-graph size $N_K$ is 10.}
  \label{fig:pgo}
   \vspace{-0.5cm}
\end{figure}

In monocular VO, continuous local BA based on the covisibility graph is essential for optimizing historical keyframes and map points, ensuring high precision and local metric consistency. This establishes the entire traditional VO framework as a reliable geometric and scale baseline in our system. However, directly leveraging this traditional VO trajectory to scale the feed-forward VGGT sub-graphs via $Sim(3)$ alignment can be numerically unstable due to the inherent scale ambiguity and gradual drift of monocular systems. Consequently, it is imperative to design an optimization mechanism to jointly refine and stabilize the scale factor.

To this end, we propose a tightly coupled hierarchical optimization backend implemented using GTSAM \cite{gtsam}. The primary stage performs traditional local BA to establish a metric foundation and ensure consistency over the local window. The secondary stage introduces a local PGO mechanism. In this stage, the wide-baseline deep learning predictions provided by VGGT act as structural constraints to locally smooth the VO poses. Together, these two stages tightly couple the traditional VO backend framework with the geometric predictions of VGGT, enhancing the accuracy of pose estimation and improving the scale consistency of dense mapping.

\textbf{Covisibility-based Local Bundle Adjustment:}
The first stage optimization performs local BA to refine the active keyframes and map points. We construct a covisibility graph and select historical keyframes with at least 50 shared landmarks with the current keyframe. The local BA minimizes the standard visual reprojection error:

\begin{equation}
\min_{\mathbf{T}, \mathbf{p}} \sum_{i \in K} \sum_{j \in P_i} \rho \left( \left\| \mathbf{z}_{i,j} - \pi(\mathbf{T}_i, \mathbf{p}_j) \right\|^2_{\bm{\Sigma}_{i,j}} \right),
\end{equation}
where $\mathbf{T}_i \in SE(3)$ is the pose of keyframe $i$, $\mathbf{p}_j$ is the 3D position of map point $j$, $\mathbf{z}_{i,j}$ is the 2D observation, $\pi(\cdot)$ is the camera projection function and $\rho(\cdot)$ is the Huber robust kernel function. The covariance matrix $\bm{\Sigma}_{i,j}$ employs the adaptive noise scale defined in Section III-B. 

To ensure stability, the poses of keyframes with low covisibility scores are fixed as priors. The local BA optimization is executed in two stages: the first stage employs a Huber kernel to down-weight outliers, followed by a Chi-square test. Observations exceeding the $\chi^2$ threshold are discarded. Subsequently, a second stage of fine-grained optimization is performed without the robust kernel to achieve maximum metric accuracy.

\textbf{Local Pose Graph Optimization:}
As illustrated in Fig.~\ref{fig:pgo}, our local PGO framework integrates two primary types of constraints: (i) scale-aware keyframe constraints derived from VGGT pose predictions, and (ii) keyframe relative pose constraints provided by the traditional VO.

To properly address the inherent scale ambiguity of the VGGT predictions, we explicitly model the scale factor $s \in \mathbb{R}^+$ as a distinct state variable. Let $\mathbf{T}_i, \mathbf{T}_j \in SE(3)$ denote the global poses of two keyframes. The initial keyframe poses are provided by the traditional VO, while the initial scale $s$ is computed via a rapid $Sim(3)$ alignment.

To formulate the VGGT pose constraint as a scale-aware relative pose factor,
we first define the raw relative pose predicted by the VGGT network as $\Delta\tilde{\mathbf{T}}_{v} = [\Delta\tilde{\mathbf{R}}, \Delta\tilde{\mathbf{t}}]$.
By explicitly incorporating the scale variable $s$, this raw prediction is scaled directly into the camera frame as $[\Delta\tilde{\mathbf{R}}, s \cdot \Delta\tilde{\mathbf{t}}]$. Meanwhile, the relative pose derived from the current state estimates is defined as $\Delta\mathbf{T}_{p} = \mathbf{T}_i^{-1} \mathbf{T}_j = [\mathbf{R}_{p}, \mathbf{t}_{p}]$. Thus, the 6-DoF residual vector $\mathbf{r}_{v} = [\mathbf{r}_\theta^T, \mathbf{r}_t^T]^T \in \mathbb{R}^6$, which decouples rotation and translation, is formulated as:

\begin{equation}
\mathbf{r}_{v}(\mathbf{T}_i, \mathbf{T}_j, s) = 
\begin{bmatrix}
\text{Log} \left( \Delta\tilde{\mathbf{R}}^T \mathbf{R}_{p} \right) \\
\mathbf{t}_{p} - s \cdot \Delta\tilde{\mathbf{t}}
\end{bmatrix},
\end{equation}
where $\text{Log}: SO(3) \rightarrow \mathbb{R}^3$ maps the rotation error to the tangent space. The Jacobian with respect to the scale state variable $s$ can be analytically derived as:

\begin{equation}
\frac{\partial \mathbf{r}_{v}}{\partial s} = \mathbf{J}_{s} = 
\begin{bmatrix}
\mathbf{0}_{3 \times 1} \\
-\Delta\tilde{\mathbf{t}}
\end{bmatrix} \in \mathbb{R}^{6 \times 1}.
\end{equation}

In each local PGO update, we jointly optimize the continuous VO trajectory and two consecutive VGGT sub-graphs connected through an overlapping keyframe. In this graph, scale consistency is primarily maintained by the classical VO trajectory, while VGGT relative-pose factors provide geometric regularization for local structure. Since VGGT does not provide pose uncertainty, we assign an empirical covariance to each relative-pose factor. For a factor from source $m \in \{\mathrm{VO}, \mathrm{VGGT}\}$, we define the noise parameters as $\sigma_t^{m} = \max\left(\sigma_{t,\min}^{m}, \lambda_t^{m} \|\mathbf{t}^{m}\|_2 \right)$ for translation and $\sigma_r^{m} = \sigma_{r,0}^{m}$ for rotation. In our implementation, we use $(\lambda_t, \sigma_{t,\min}, \sigma_{r,0}) = (0.02, 0.02, 0.01\text{ rad})$ for VO factors and $(0.10, 0.10, 0.02\text{ rad})$ for VGGT factors. These larger noise parameters assign lower confidence to VGGT factors, so that they act as auxiliary geometric constraints. The relative translation of each VGGT factor is associated with a local scale variable initialized by local trajectory alignment.

%% file: article/experiments.tex
\subsection{Experimental Setup}
We evaluate all methods on the same server equipped with an Intel Xeon Gold 6342 CPU and an NVIDIA RTX 4090 GPU with 24GB VRAM to ensure a consistent evaluation. Experiments are conducted on EuRoC MAV~\cite{euroc} and TUM RGB-D for indoor scenes, and on KITTI~\cite{kitti} for large-scale outdoor scenarios. We report ATE on EuRoC MAV and TUM RGB-D, and the relative pose error metrics on KITTI. For monocular methods, all trajectory-based metrics are computed after 7-DoF Sim(3) alignment with the ground-truth trajectory. All results are reported as mean $\pm$ standard deviation over five runs, where ``$<0.01$'' denotes standard deviation below the displayed precision in the table. For methods that provide official loop-closure options, we use their official source-code interfaces to enable or disable these modules. In the tables, ``w/o LC'' denotes odometry-only results, while ``w/ LC'' denotes results with loop closure when available. We also include a method, \textit{VO+DA3}, in Tabs.~\ref{tab:euroc_eval}--\ref{tab:kitti_00_05} to evaluate a simpler design that directly combines the classical VO pose with Depth-Anything 3.

\subsection{Dataset Evaluation}

\textbf{Indoor Evaluation on EuRoC MAV and TUM RGB-D:}
We evaluate indoor trajectory accuracy on the EuRoC MAV and TUM RGB-D datasets, with the ATE results summarized in Tables~\ref{tab:euroc_eval} and~\ref{tab:tum_rgbd_eval}. TUM RGB-D results are reported in centimeters, while EuRoC results are reported in meters.

\definecolor{best}{HTML}{A7D0A7} 
\definecolor{second}{HTML}{E2EFE2} 

\input{tables/Table_EuRoC}

\definecolor{best}{HTML}{A7D0A7} 
\definecolor{second}{HTML}{E2EFE2} 

\begin{figure}[!t]
  \centering
  \includegraphics[width=1.0\linewidth]{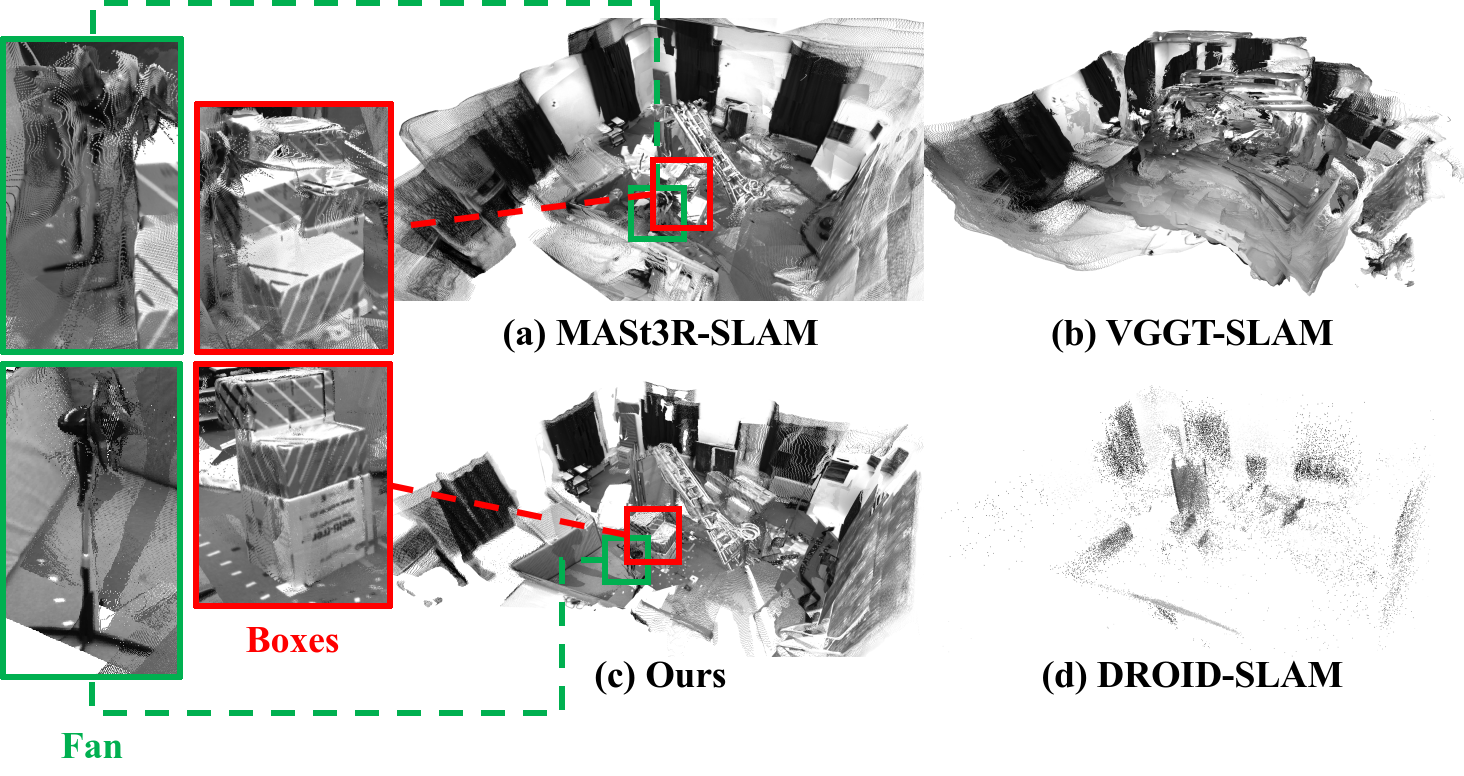}
   \vspace{-20pt}
  \caption{Qualitative comparison of dense 3D indoor reconstructions on the EuRoC V202 sequence. Prominently, (b) VGGT-SLAM exhibits noticeable reconstruction inconsistencies and structural distortion, while (d) DROID-SLAM yields a highly sparse map caused by its aggressive confidence filtering and down-sample strategy. In contrast, both (a) MASt3R-SLAM and our proposed method (c) maintain global structural integrity. As highlighted in the enlarged views, our method preserves richer local structural details and more consistent geometry in the selected regions compared with (a).} 
  \label{fig:euroc_mapping}
  \vspace{-0.5cm}
\end{figure}

\input{tables/Table_TUM_RGB-D}
On EuRoC MAV, HyVGGT-VO achieves an average ATE of 0.20 m in the odometry-only setting, close to DROID-SLAM with an average ATE of 0.19 m. Compared with feed-forward baselines, HyVGGT-VO reduces the average ATE by approximately 85.0\% over VGGT-SLAM (\textit{Sim3}) and 90.5\% over VGGT-SLAM 2.0. ORB-SLAM3 achieves lower errors than the feed-forward baselines and HyVGGT-VO on most sequences. However, both ORB-SLAM3 and DSO fail on V203, whereas HyVGGT-VO successfully completes the sequence. Nevertheless, V203 remains a challenging sequence for our system and is the main contributor to the remaining gap with DROID-SLAM. In this sequence, the KLT--PnP frontend becomes unstable under fast motion and motion blur, triggering the VGGT recovery strategy. This recovery maintains trajectory continuity, but degraded VGGT predictions on grayscale blurred images and pose drift still lead to lower accuracy than DROID-SLAM. The dense mapping result in Fig.~\ref{fig:euroc_mapping} is generated by back-projecting the scaled VGGT depth maps with the optimized camera poses, followed by VGGT confidence-based depth filtering and point-cloud concatenation.

Under the odometry-only setting, HyVGGT-VO achieves an average ATE of $4.38$ cm over all nine TUM RGB-D sequences, which is the lowest average among the evaluated methods. DROID-SLAM and VGGT-SLAM 2.0 obtain average ATEs of $9.88$ cm and $5.49$ cm, respectively. For the traditional VO methods, ORB-SLAM3 and DSO fail on most sequences, although ORB-SLAM3 achieves lower errors on the remaining sequences. Overall, HyVGGT-VO provides more robust and competitive pose estimation across the benchmark while additionally supporting dense reconstruction.

\textbf{Outdoor Evaluation on KITTI Dataset:}
We further evaluate HyVGGT-VO on the KITTI benchmark, which contains high-speed vehicle motion and large viewpoint changes. The results for sequences 00--05 are reported in Tab.~\ref{tab:kitti_00_05} using the KITTI relative pose error metrics.

In the odometry-only setting on Seq.~00--05, HyVGGT-VO achieves an average $t_{\rm rel}$ of $15.43\%$ and the lowest average $r_{\rm rel}$ of $0.20$ deg/100m. Among the evaluated VGGT-based methods, VGGT-Long and HyVGGT-VO achieve the lowest and second-lowest average $t_{\rm rel}$, respectively; however, VGGT-Long has a higher average $r_{\rm rel}$ than HyVGGT-VO ($1.54$ vs. $0.20$ deg/100m). The traditional ORB-SLAM3 baseline is highly competitive, with lower translation and rotation errors than the evaluated feed-forward methods and HyVGGT-VO on Seq.~02 and Seq.~03. Its main degradation occurs on Seq.~01, where it suffers severe translational drift with a $t_{\rm rel}$ of $76.45\%$, whereas HyVGGT-VO achieves the lowest error of $3.65\%$.

On Seq.~06--10, the traditional methods are more competitive in the odometry-only setting. ORB-SLAM3 achieves the lowest average $t_{\rm rel}$ of $12.60\%$, followed by VGGT-Long at $13.67\%$, while HyVGGT-VO obtains an average $t_{\rm rel}$ of $20.12\%$. In terms of rotation, HyVGGT-VO maintains a low average $r_{\rm rel}$ of $0.29$ deg/100m, comparable to ORB-SLAM3 ($0.28$ deg/100m), although DSO achieves the lowest value of $0.26$ deg/100m. Compared with the evaluated VGGT-based methods, HyVGGT-VO achieves lower translational errors than VGGT-SLAM (Sim3) and VGGT-SLAM 2.0 on Seq.~07, Seq.~09, and Seq.~10. Nevertheless, ORB-SLAM3 consistently achieves lower translational errors on Seq.~06--10, indicating that classical geometric odometry remains highly competitive on these sequences.

\begin{figure}[!t]
  \centering
  \includegraphics[width=1.0\linewidth]{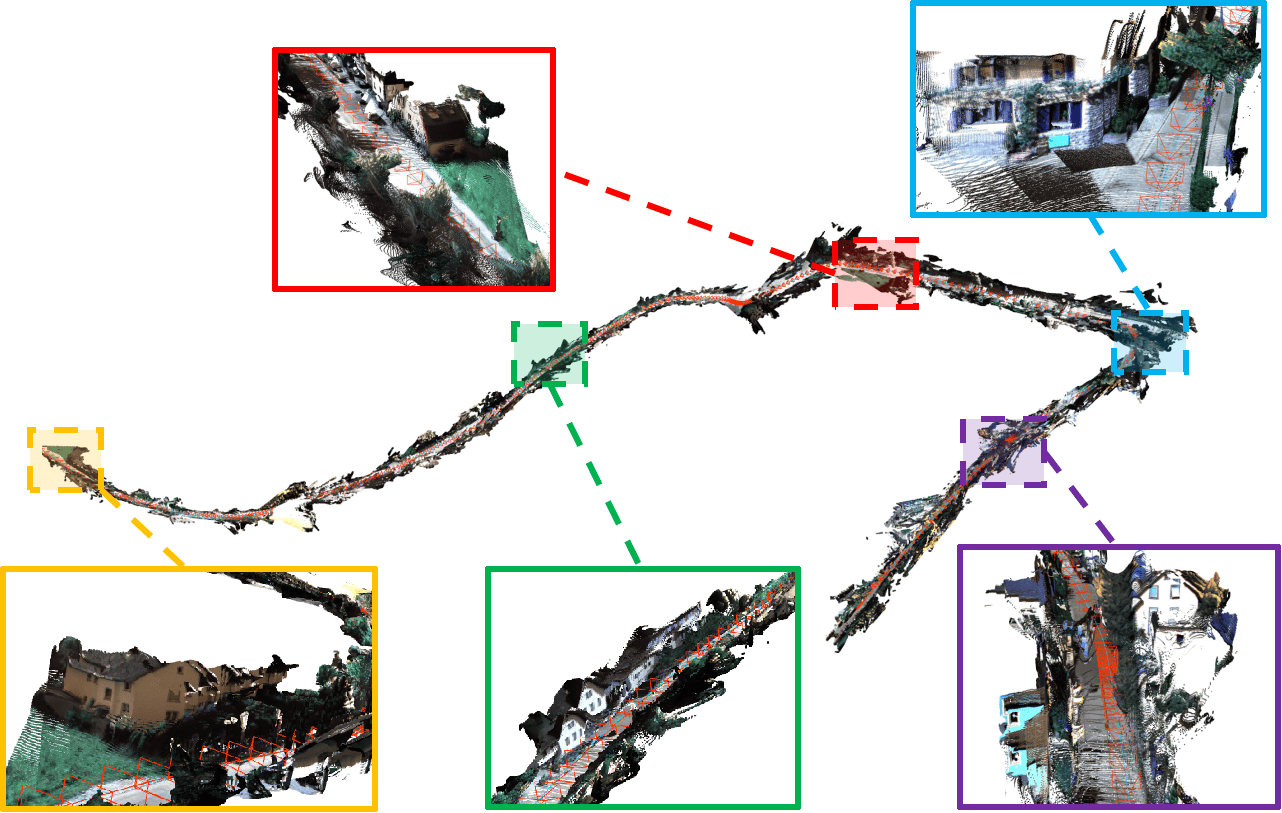}
  \vspace{-0.6cm}
  \caption{Large-scale outdoor reconstruction over a 920 m trajectory with approximately 20.1 million 3D points before voxel downsampling.}
  \label{fig:kitti_mapping}
  \vspace{-0.5cm}
\end{figure}

The dense reconstruction on KITTI Seq.~10 is shown in Fig.~\ref{fig:kitti_mapping}, demonstrating that our system maintains consistent dense mapping in outdoor scenes.

\definecolor{best}{HTML}{A7D0A7} 
\definecolor{second}{HTML}{E2EFE2} 
\input{tables/Table_KITTI_00-05}
\input{tables/Table_KITTI_06-10}

\subsection{Ablation Study}
\textbf{Adaptive Hybrid Frontend Ablation:}
To validate the effectiveness of our adaptive hybrid frontend, we evaluate our system on the EuRoC and KITTI datasets using a KLT-only baseline, where the VGGT tracking head is explicitly disabled. As shown in Tab. \ref{table:ablation_frontend}, the KLT-only tracker completely fails on several difficult sequences when confronted with drastic illumination changes (EuRoC V103), rapid FOV shifts (KITTI 01), repetitive textures (KITTI 00), and severe motion blur (EuRoC V203). In contrast, the proposed hybrid frontend maintains stable pose estimation in these cases.

\begin{table}[!t]
\centering
\caption{Frontend ablation on challenging sequences. KITTI uses $t_{\mathrm{rel}}$~(\%)/$r_{\mathrm{rel}}$~(deg/100m), and EuRoC uses ATE RMSE (m).}
\vspace{-0.2cm}
\label{table:ablation_frontend}
\begin{threeparttable}
\small
\renewcommand{\arraystretch}{1.05}
\begin{tabular*}{\columnwidth}{@{\extracolsep{\fill}}lcccc@{}}
\toprule
\multirow{2}{*}{\textbf{Method}} 
& \multicolumn{2}{c}{\textbf{KITTI}} 
& \multicolumn{2}{c}{\textbf{EuRoC}} \\
\cmidrule(lr){2-3} \cmidrule(l){4-5}
& \textbf{00} & \textbf{01} & \textbf{V103} & \textbf{V203} \\
\midrule
KLT-only & $\times$ & $\times$ & $\times$ & $\times$ \\
Hybrid Frontend & 33.24/0.27 & 3.65/0.13 & 0.23 & 0.59 \\
\bottomrule
\end{tabular*}
\end{threeparttable}
\vspace{-0.5cm}
\end{table}

\begin{figure}[!t]
  \centering
  \includegraphics[width=1.0\linewidth]{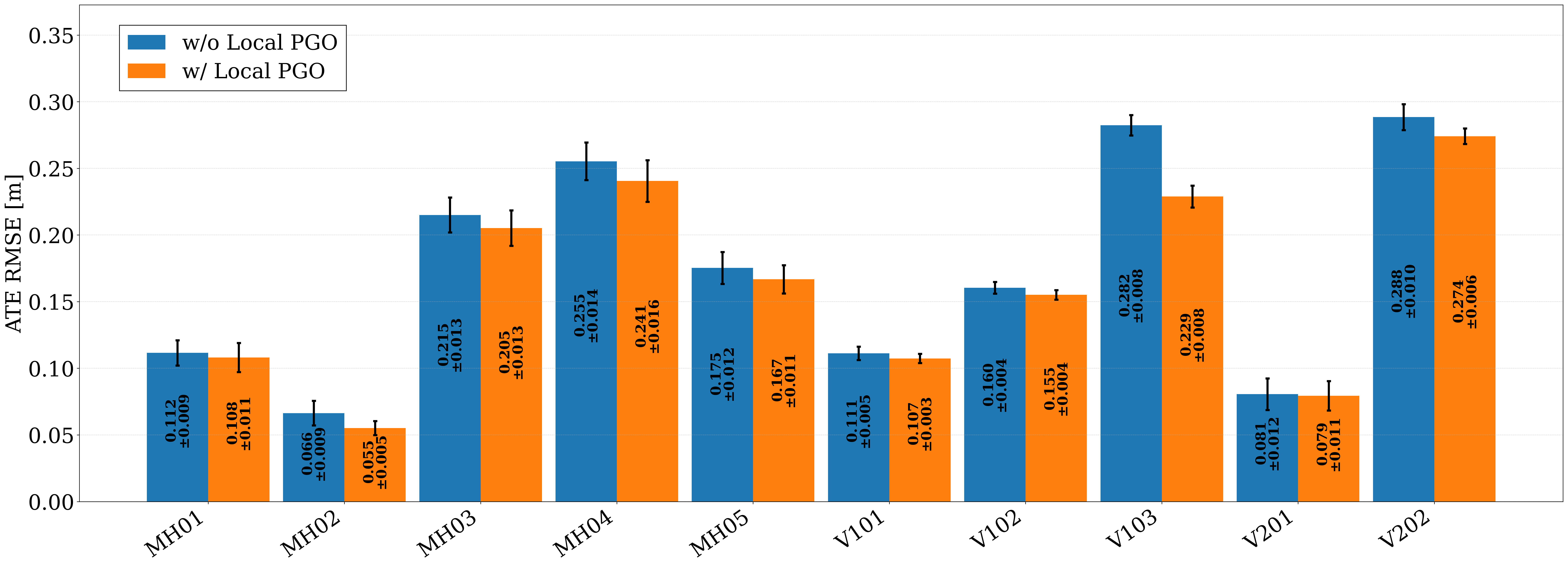}
   \vspace{-20pt}
  \caption{Quantitative ablation of the local PGO module on EuRoC sequences. Bars report the mean ATE over multiple independent runs, and error bars denote the standard deviation. V203 is excluded because the classical VO fails to maintain valid pose estimation on this sequence, so the local PGO module is rarely used.}
  \label{fig:ablation_pgo}
   \vspace{-0.5cm}
\end{figure}

\textbf{Local PGO Ablation:}
To validate the effectiveness of the proposed local PGO module, we conduct ablation experiments on the EuRoC dataset by disabling the local PGO module as the baseline. As shown in Fig.~\ref{fig:ablation_pgo}, when the classical VO frontend performs well, it provides accurate pose estimates, and the improvement brought by local PGO is therefore less evident. In this case, the low-weight VGGT pose factors have little influence. In contrast, on V103, where the classical VO trajectory exhibits relatively large drift, the VGGT pose factors provide more robust constraints and play a more significant role in improving trajectory consistency, thereby enhancing the robustness of the system.

\subsection{Real-time Performance and Efficiency}
We evaluate runtime on the EuRoC MH01--MH05 sequences using two platforms: a workstation with an Intel Core i9-14900KF CPU, 64GB RAM, and an NVIDIA RTX 4090 GPU with 48GB VRAM, and a laptop with an AMD Ryzen 9 7940HX CPU, 32GB RAM, and an NVIDIA RTX 4060 Laptop GPU with 8GB VRAM. The workstation is used for full-method runtime comparison, while the laptop evaluates deployability under a strict memory budget. We separately report the pose-estimation output rate and dense-mapping update rate.

\input{tables/Table_speed_compare}

As shown in Tab.~\ref{table:system_speed}, HyVGGT-VO uses the lowest GPU memory among the compared systems on both platforms. On the RTX 4090 workstation, it provides pose estimates at $34.22$ Hz, faster than the evaluated feed-forward methods but slower than DROID-SLAM, and updates dense maps at $21.01$ Hz, comparable to the VGGT-based methods. On the RTX 4060 Laptop GPU with 8GB VRAM, most baselines exceed the available memory, whereas HyVGGT-VO is executable with $16.07$ Hz pose output and $3.17$ Hz dense-mapping updates, compared with approximately $3.30$ Hz for both outputs from VGGT-SLAM 2.0. These results show that HyVGGT-VO balances continuous pose output, feed-forward dense mapping, and GPU memory efficiency.

Since invoking the VGGT model introduces additional inference cost, we further evaluate its runtime impact on the RTX 4060 Laptop platform by reporting the per-sequence pose-output rate and VGGT invocation frequency on the EuRoC MH sequences in Tab.~\ref{tab:recovery_fps}. On MH01--MH03, the frontend relies primarily on classical VO, with VGGT invoked only once or twice. More frequent VGGT invocation is observed on MH04 and MH05 due to the more challenging conditions, such as low illumination and fast motion.

In addition, Tab.~\ref{tab:timing_analysis} further reports the execution times of the classical frontend, backend optimization, and VGGT inference on the same platform.

\input{tables/Table_recover}
\input{tables/Table_runtime_analysis}

%% file: tables/Table_EuRoC.tex
%
\providecommand{\pmv}[2]{\ensuremath{#1\!\pm\!#2}}
\providecolor{best}{RGB}{221,235,247}
\providecolor{second}{RGB}{226,240,217}

\begin{table*}[!t]
\centering
\caption{ATE(m) Results on the EuRoC MAV dataset}
\vspace{-0.3cm}
\label{tab:euroc_eval}
\small
\setlength{\tabcolsep}{1.4pt}
\renewcommand{\arraystretch}{1.05}
\resizebox{\textwidth}{!}{%
\begin{tabular}{@{}llcccccccccccc@{}}
\toprule
\multirow{2}{*}{Mode} & \multirow{2}{*}{Method}
& \multicolumn{5}{c}{\textbf{Machine Hall (MH)}}
& \multicolumn{6}{c}{\textbf{Vicon Room (V)}}
& \multirow{2}{*}{Avg.} \\
\cmidrule(lr){3-7}\cmidrule(lr){8-13}
& & 01 & 02 & 03 & 04 & 05 & 101 & 102 & 103 & 201 & 202 & 203 & \\
\midrule
\multirow{7}{*}{w/ LC}
& ORB-SLAM3
& \pmv{0.04}{{<}0.01}
& \pmv{0.04}{0.01}
& \cellcolor{second}\pmv{0.04}{{<}0.01}
& \cellcolor{second}\pmv{0.07}{0.04}
& \cellcolor{second}\pmv{0.06}{0.01}
& \pmv{0.09}{{<}0.01}
& \pmv{0.06}{{<}0.01}
& \pmv{0.11}{0.05}
& \pmv{0.20}{0.05}
& \pmv{0.08}{0.05}
& --
& -- \\

& DROID-SLAM
& \cellcolor{best}\pmv{0.01}{{<}0.01}
& \cellcolor{best}\pmv{0.01}{{<}0.01}
& \cellcolor{best}\pmv{0.02}{{<}0.01}
& \cellcolor{best}\pmv{0.04}{{<}0.01}
& \cellcolor{best}\pmv{0.04}{{<}0.01}
& \cellcolor{best}\pmv{0.04}{{<}0.01}
& \cellcolor{best}\pmv{0.01}{{<}0.01}
& \cellcolor{best}\pmv{0.02}{{<}0.01}
& \cellcolor{best}\pmv{0.02}{{<}0.01}
& \cellcolor{best}\pmv{0.01}{{<}0.01}
& \cellcolor{best}\pmv{0.01}{{<}0.01}
& \cellcolor{best}0.02 \\

& MASt3R-SLAM
& \cellcolor{second}\pmv{0.02}{{<}0.01}
& \cellcolor{second}\pmv{0.02}{{<}0.01}
& \pmv{0.06}{{<}0.01}
& \pmv{0.11}{{<}0.01}
& \pmv{0.07}{{<}0.01}
& \cellcolor{second}\pmv{0.04}{{<}0.01}
& \cellcolor{second}\pmv{0.02}{{<}0.01}
& \cellcolor{second}\pmv{0.03}{{<}0.01}
& \cellcolor{second}\pmv{0.02}{{<}0.01}
& \cellcolor{second}\pmv{0.03}{{<}0.01}
& \cellcolor{second}\pmv{0.04}{{<}0.01}
& \cellcolor{second}0.04 \\

& ViSTA-SLAM
& \pmv{2.52}{0.02}
& \pmv{1.11}{0.05}
& \pmv{1.49}{{<}0.01}
& \pmv{3.97}{0.04}
& \pmv{4.06}{0.01}
& \pmv{0.90}{{<}0.01}
& \pmv{1.51}{0.01}
& \pmv{1.43}{{<}0.01}
& \pmv{1.01}{{<}0.01}
& \pmv{1.65}{0.01}
& \pmv{1.86}{0.01}
& 1.96 \\

& VGGT-SLAM(Sim3)
& \pmv{1.05}{0.04}
& \pmv{2.93}{0.04}
& \pmv{2.30}{{<}0.01}
& \pmv{3.46}{0.01}
& \pmv{3.54}{{<}0.01}
& \pmv{0.68}{{<}0.01}
& \pmv{1.67}{{<}0.01}
& \pmv{0.59}{{<}0.01}
& \pmv{0.53}{{<}0.01}
& \pmv{0.94}{{<}0.01}
& \pmv{1.16}{{<}0.01}
& 1.71 \\

& VGGT-SLAM(SL4)
& \pmv{2.63}{{<}0.01}
& \pmv{1.76}{{<}0.01}
& \pmv{2.51}{0.16}
& \pmv{6.74}{0.14}
& \pmv{3.03}{0.12}
& \pmv{0.71}{0.13}
& --
& --
& \pmv{0.43}{0.01}
& --
& --
& -- \\

& VGGT-SLAM2.0
& \pmv{0.76}{{<}0.01}
& \pmv{0.93}{{<}0.01}
& \pmv{1.63}{{<}0.01}
& \pmv{1.88}{{<}0.01}
& \pmv{2.76}{{<}0.01}
& \pmv{0.57}{{<}0.01}
& \pmv{0.56}{{<}0.01}
& \pmv{0.54}{{<}0.01}
& \pmv{0.40}{{<}0.01}
& \pmv{0.57}{{<}0.01}
& \pmv{0.73}{{<}0.01}
& 1.03 \\
\midrule
\multirow{10}{*}{\makecell[c]{w/o LC\\(Odom.)}}
& DSO
& \cellcolor{second}\pmv{0.05}{0.01}
& \pmv{0.04}{{<}0.01}
& \cellcolor{second}\pmv{0.17}{0.02}
& \cellcolor{second}\pmv{0.18}{0.01}
& \cellcolor{best}\pmv{0.11}{0.01}
& \pmv{0.16}{0.08}
& \pmv{0.80}{0.72}
& \pmv{0.92}{0.22}
& \pmv{0.07}{0.01}
& \cellcolor{second}\pmv{0.11}{0.03}
& --
& -- \\

& ORB-SLAM3
& \cellcolor{best}\pmv{0.04}{{<}0.01}
& \cellcolor{best}\pmv{0.03}{{<}0.01}
& \cellcolor{best}\pmv{0.04}{{<}0.01}
& \cellcolor{best}\pmv{0.08}{0.02}
& \pmv{0.25}{0.05}
& \cellcolor{best}\pmv{0.09}{{<}0.01}
& \cellcolor{best}\pmv{0.10}{0.01}
& \pmv{0.48}{0.08}
& \pmv{0.27}{0.06}
& \cellcolor{best}\pmv{0.07}{0.05}
& --
& -- \\

& DROID-SLAM
& \pmv{0.16}{0.01}
& \pmv{0.12}{0.02}
& \pmv{0.24}{0.01}
& \pmv{0.40}{{<}0.01}
& \pmv{0.27}{{<}0.01}
& \cellcolor{second}\pmv{0.10}{{<}0.01}
& \pmv{0.17}{0.01}
& \cellcolor{second}\pmv{0.16}{0.01}
& \pmv{0.10}{{<}0.01}
& \pmv{0.12}{0.01}
& \cellcolor{best}\pmv{0.20}{{<}0.01}
& \cellcolor{best}0.19 \\

& MASt3R-SLAM
& \pmv{0.22}{{<}0.01}
& \pmv{0.59}{{<}0.01}
& \pmv{0.34}{{<}0.01}
& \pmv{2.10}{{<}0.01}
& \pmv{0.73}{{<}0.01}
& \pmv{0.12}{{<}0.01}
& \cellcolor{second}\pmv{0.15}{{<}0.01}
& \cellcolor{best}\pmv{0.13}{{<}0.01}
& \pmv{0.11}{{<}0.01}
& \pmv{0.35}{{<}0.01}
& \pmv{0.64}{{<}0.01}
& 0.50 \\

& ViSTA-SLAM
& \pmv{2.54}{0.04}
& \pmv{1.49}{0.02}
& \pmv{2.22}{0.04}
& \pmv{4.08}{0.07}
& \pmv{4.48}{0.09}
& \pmv{1.24}{0.05}
& \pmv{1.74}{{<}0.01}
& \pmv{1.51}{0.07}
& \pmv{1.01}{{<}0.01}
& \pmv{1.83}{0.01}
& \pmv{1.90}{0.03}
& 2.19 \\

& VGGT-SLAM(Sim3)
& \pmv{0.71}{{<}0.01}
& \pmv{0.95}{{<}0.01}
& \pmv{1.82}{{<}0.01}
& \pmv{3.49}{{<}0.01}
& \pmv{3.11}{{<}0.01}
& \pmv{0.68}{{<}0.01}
& \pmv{0.69}{{<}0.01}
& \pmv{0.59}{{<}0.01}
& \pmv{0.53}{{<}0.01}
& \pmv{0.94}{{<}0.01}
& \pmv{1.16}{{<}0.01}
& 1.33 \\

& VGGT-SLAM(SL4)
& \pmv{4.86}{0.01}
& \pmv{3.90}{0.01}
& \pmv{2.86}{0.03}
& \pmv{6.37}{{<}0.01}
& \pmv{4.68}{0.02}
& \pmv{1.38}{{<}0.01}
& \pmv{1.54}{0.01}
& \pmv{1.29}{{<}0.01}
& \pmv{2.10}{{<}0.01}
& \pmv{1.82}{0.01}
& \pmv{1.96}{0.01}
& 2.98 \\

& VGGT-SLAM2.0
& \pmv{1.64}{{<}0.01}
& \pmv{1.00}{{<}0.01}
& \pmv{2.41}{{<}0.01}
& \pmv{4.77}{{<}0.01}
& \pmv{3.84}{{<}0.01}
& \pmv{1.03}{{<}0.01}
& \pmv{1.93}{{<}0.01}
& \pmv{1.40}{{<}0.01}
& \pmv{1.16}{{<}0.01}
& \pmv{1.95}{{<}0.01}
& \pmv{2.11}{{<}0.01}
& 2.11 \\

& VO+DA3
& \pmv{0.11}{0.03}
& \pmv{0.06}{0.02}
& \pmv{0.31}{0.01}
& \pmv{0.25}{0.03}
& \pmv{0.20}{0.06}
& \pmv{0.10}{0.01}
& \pmv{0.16}{0.02}
& --
& \cellcolor{second}\pmv{0.07}{0.02}
& \pmv{0.32}{0.03}
& --
& -- \\

& \textbf{Ours}
& \pmv{0.10}{0.01}
& \cellcolor{second}\pmv{0.04}{0.01}
& \pmv{0.24}{0.03}
& \pmv{0.24}{0.03}
& \cellcolor{second}\pmv{0.17}{0.01}
& \cellcolor{second}\pmv{0.10}{0.01}
& \pmv{0.16}{0.02}
& \pmv{0.23}{0.01}
& \cellcolor{best}\pmv{0.07}{0.01}
& \pmv{0.27}{0.01}
& \cellcolor{second}\pmv{0.59}{0.07}
& \cellcolor{second}0.20 \\
\bottomrule
\end{tabular}%
}
\vspace{-1mm}
\end{table*}

%% file: tables/Table_TUM_RGB-D.tex
%
\providecommand{\pmv}[2]{\ensuremath{#1\!\pm\!#2}}
\providecolor{best}{RGB}{221,235,247}
\providecolor{second}{RGB}{226,240,217}

\begin{table*}[!t]
\centering
\caption{ATE(cm) on the TUM RGB-D dataset}
\vspace{-0.3cm}
\label{tab:tum_rgbd_eval}
\small
\setlength{\tabcolsep}{1.5pt}
\renewcommand{\arraystretch}{1.00}

\resizebox{\textwidth}{!}{%
\begin{tabular}{@{}llcccccccccc@{}}
\toprule
Mode & Method & 360 & desk & desk2 & floor & plant & room & rpy & teddy & xyz & Avg. \\
\midrule
\multirow{7}{*}{w/ LC}
& ORB-SLAM3
& -- & \pmv{1.70}{0.32} & -- & -- & \pmv{3.40}{1.12} & -- & -- & -- & \cellcolor{best}\pmv{0.90}{{<}0.01} & -- \\
& DROID-SLAM
& \pmv{7.14}{0.14} & \cellcolor{second}\pmv{1.64}{0.01} & \cellcolor{second}\pmv{2.64}{{<}0.01} & \cellcolor{best}\pmv{2.14}{{<}0.01} & \cellcolor{best}\pmv{1.37}{0.01} & \cellcolor{best}\pmv{4.96}{{<}0.01} & \cellcolor{best}\pmv{2.28}{0.01} & \cellcolor{best}\pmv{3.45}{0.02} & \pmv{1.04}{{<}0.01} & \cellcolor{best}2.96 \\
& MASt3R-SLAM
& \cellcolor{best}\pmv{4.91}{{<}0.01} & \cellcolor{best}\pmv{1.62}{{<}0.01} & \cellcolor{best}\pmv{2.43}{{<}0.01} & \cellcolor{second}\pmv{2.55}{{<}0.01} & \cellcolor{second}\pmv{2.43}{{<}0.01} & \pmv{6.11}{{<}0.01} & \pmv{2.73}{{<}0.01} & \pmv{4.18}{{<}0.01} & \cellcolor{second}\pmv{0.91}{{<}0.01} & \cellcolor{second}3.10 \\
& ViSTA-SLAM
& \pmv{10.42}{0.02} & \pmv{3.00}{{<}0.01} & \pmv{2.97}{0.02} & \pmv{6.99}{0.01} & \pmv{5.22}{0.02} & \pmv{6.74}{0.04} & \cellcolor{second}\pmv{2.30}{{<}0.01} & \pmv{7.99}{0.01} & \pmv{1.49}{0.01} & 5.24 \\
& \makecell[l]{VGGT-SLAM (Sim3)}
& \pmv{5.42}{{<}0.01} & \pmv{2.52}{{<}0.01} & \pmv{3.79}{{<}0.01} & \pmv{4.77}{{<}0.01} & \pmv{2.68}{{<}0.01} & \cellcolor{second}\pmv{5.53}{{<}0.01} & \pmv{2.74}{{<}0.01} & \pmv{3.81}{{<}0.01} & \pmv{1.84}{{<}0.01} & 3.68 \\
& \makecell[l]{VGGT-SLAM (SL4)}
& \pmv{12.38}{4.85} & \pmv{2.99}{0.37} & \pmv{8.13}{2.97} & \pmv{11.82}{2.75} & \pmv{2.64}{0.42} & \pmv{10.74}{2.56} & \pmv{3.83}{1.33} & \pmv{7.58}{4.17} & \pmv{4.44}{3.36} & 7.17 \\
& VGGT-SLAM 2.0
& \cellcolor{second}\pmv{5.00}{{<}0.01} & \pmv{2.50}{{<}0.01} & \pmv{2.90}{{<}0.01} & \pmv{10.20}{{<}0.01} & \pmv{2.60}{{<}0.01} & \pmv{6.30}{{<}0.01} & \pmv{2.60}{{<}0.01} & \cellcolor{second}\pmv{3.80}{{<}0.01} & \pmv{1.40}{{<}0.01} & 4.14 \\
\midrule
\multirow{10}{*}{\makecell[c]{w/o LC\\(Odom.)}}
& DSO
& \pmv{14.88}{4.09} & \pmv{18.97}{10.69} & \pmv{71.02}{18.16} & \pmv{23.09}{5.04} & \pmv{45.82}{10.82} & \pmv{57.81}{25.88} & \pmv{57.81}{25.88} & -- & \pmv{6.19}{0.94} & -- \\
& ORB-SLAM3
& -- & \cellcolor{best}\pmv{1.79}{0.32} & -- & -- & \pmv{3.42}{1.59} & -- & -- & -- & \cellcolor{best}\pmv{0.93}{{<}0.01} & -- \\
& DROID-SLAM
& \pmv{16.12}{0.46} & \pmv{2.81}{0.07} & \pmv{9.95}{0.06} & \cellcolor{best}\pmv{3.39}{0.02} & \cellcolor{best}\pmv{2.81}{0.25} & \pmv{32.70}{0.32} & \pmv{2.84}{0.03} & \pmv{16.92}{0.19} & \cellcolor{second}\pmv{1.39}{1.13} & 9.88 \\
\addlinespace[0.20em]
& MASt3R-SLAM
& \pmv{8.33}{{<}0.01} & \pmv{6.99}{{<}0.01} & \pmv{9.76}{{<}0.01} & \pmv{5.55}{{<}0.01} & \pmv{5.46}{{<}0.01} & \cellcolor{second}\pmv{8.23}{{<}0.01} & \pmv{4.10}{{<}0.01} & \pmv{29.28}{{<}0.01} & \pmv{1.65}{{<}0.01} & 8.82 \\
& ViSTA-SLAM
& \pmv{13.26}{{<}0.01} & \pmv{7.32}{{<}0.01} & \pmv{6.96}{{<}0.01} & \pmv{11.23}{{<}0.01} & \pmv{22.49}{{<}0.01} & \pmv{13.42}{{<}0.01} & \pmv{3.59}{{<}0.01} & \pmv{21.82}{{<}0.01} & \pmv{2.56}{0.03} & 11.41 \\
& \makecell[l]{VGGT-SLAM (Sim3)}
& \pmv{7.56}{{<}0.01} & \pmv{2.80}{{<}0.01} & \cellcolor{second}\pmv{6.58}{{<}0.01} & \pmv{7.43}{{<}0.01} & \pmv{3.94}{{<}0.01} & \pmv{10.53}{{<}0.01} & \cellcolor{second}\pmv{2.77}{{<}0.01} & \cellcolor{second}\pmv{6.20}{{<}0.01} & \pmv{1.85}{{<}0.01} & 5.52 \\
& \makecell[l]{VGGT-SLAM (SL4)}
& \pmv{16.92}{2.05} & \pmv{3.16}{0.75} & \pmv{25.09}{16.35} & \pmv{35.54}{15.64} & \cellcolor{second}\pmv{3.25}{0.20} & \pmv{57.09}{38.23} & \pmv{5.35}{0.16} & \pmv{7.39}{1.14} & \pmv{2.95}{1.17} & 17.42 \\
& VGGT-SLAM 2.0
& \cellcolor{second}\pmv{6.54}{{<}0.01} & \pmv{3.43}{{<}0.01} & \pmv{10.01}{{<}0.01} & \cellcolor{second}\pmv{4.52}{{<}0.01} & \pmv{4.01}{{<}0.01} & \pmv{9.18}{{<}0.01} & \pmv{2.78}{{<}0.01} & \pmv{7.10}{{<}0.01} & \pmv{1.86}{{<}0.01} & \cellcolor{second}5.49 \\
\addlinespace[0.20em]
& VO+DA3
& -- & \pmv{4.50}{0.80} & -- & -- & \pmv{40.60}{7.30} & -- & -- & \pmv{84.50}{10.60} & \pmv{3.30}{0.30} & -- \\
& \textbf{Ours}
& \cellcolor{best}\pmv{5.32}{0.18} & \cellcolor{second}\pmv{2.53}{0.04} & \cellcolor{best}\pmv{5.43}{0.23} & \pmv{5.10}{0.50} & \pmv{3.58}{0.02} & \cellcolor{best}\pmv{8.18}{0.11} & \cellcolor{best}\pmv{2.66}{0.02} & \cellcolor{best}\pmv{4.99}{0.06} & \pmv{1.59}{0.03} & \cellcolor{best}4.38 \\
\bottomrule
\end{tabular}%
}

\vspace{-0.3cm}
\end{table*}

%% file: tables/Table_KITTI_00-05.tex
%
\providecommand{\pmv}[2]{\ensuremath{#1\!\pm\!#2}}
\definecolor{best}{HTML}{A7D0A7}
\definecolor{second}{HTML}{E2EFE2}

\begin{table*}[!t]
\centering
\caption{KITTI odometry results on Seq. 00--05 using $t_{\rm rel}$ (\%) and $r_{\rm rel}$ (deg/100m).}
\vspace{-0.3cm}
\label{tab:kitti_00_05}
\small
\begingroup
\setlength{\tabcolsep}{1.1pt}
\renewcommand{\arraystretch}{1.05}

\resizebox{\textwidth}{!}{%
\begin{tabular}{@{}llcccccccccccccc@{}}
\toprule
\multirow{2}{*}{Mode} & \multirow{2}{*}{Method}
& \multicolumn{2}{c}{\textbf{00}}
& \multicolumn{2}{c}{\textbf{01}}
& \multicolumn{2}{c}{\textbf{02}}
& \multicolumn{2}{c}{\textbf{03}}
& \multicolumn{2}{c}{\textbf{04}}
& \multicolumn{2}{c}{\textbf{05}}
& \multicolumn{2}{c}{\textbf{Avg.}} \\
\cmidrule(lr){3-4}\cmidrule(lr){5-6}\cmidrule(lr){7-8}\cmidrule(lr){9-10}\cmidrule(lr){11-12}\cmidrule(lr){13-14}\cmidrule(lr){15-16}
& & $t_{\rm rel}$ & $r_{\rm rel}$ & $t_{\rm rel}$ & $r_{\rm rel}$ & $t_{\rm rel}$ & $r_{\rm rel}$ & $t_{\rm rel}$ & $r_{\rm rel}$ & $t_{\rm rel}$ & $r_{\rm rel}$ & $t_{\rm rel}$ & $r_{\rm rel}$ & $t_{\rm rel}$ & $r_{\rm rel}$ \\
\midrule
\multirow{7}{*}{w/ LC}
& ORB-SLAM3
& \cellcolor{best}\pmv{3.09}{1.27} & \cellcolor{second}\pmv{0.54}{0.17}
& \pmv{87.29}{8.61} & \cellcolor{second}\pmv{0.54}{0.10}
& \cellcolor{best}\pmv{4.37}{0.33} & \cellcolor{second}\pmv{0.32}{0.06}
& \cellcolor{best}\pmv{0.91}{0.15} & \cellcolor{best}\pmv{0.17}{0.01}
& \cellcolor{best}\pmv{0.75}{0.15} & \cellcolor{best}\pmv{0.16}{0.03}
& \cellcolor{second}\pmv{4.39}{1.36} & \cellcolor{second}\pmv{0.42}{0.12}
& \cellcolor{second}16.80 & \cellcolor{second}0.36 \\

& DROID-SLAM
& \pmv{26.49}{0.01} & \cellcolor{best}\pmv{0.29}{{<}0.01}
& \cellcolor{second}\pmv{20.83}{1.05} & \cellcolor{best}\pmv{0.14}{0.01}
& \pmv{32.17}{0.85} & \cellcolor{best}\pmv{0.24}{0.01}
& \cellcolor{second}\pmv{1.36}{{<}0.01} & \cellcolor{second}\pmv{0.25}{{<}0.01}
& \cellcolor{second}\pmv{1.58}{{<}0.01} & \cellcolor{second}\pmv{0.35}{{<}0.01}
& \pmv{22.65}{0.01} & \cellcolor{best}\pmv{0.26}{{<}0.01}
& 17.51 & \cellcolor{best}0.26 \\

& MASt3R-SLAM
& -- & --
& -- & --
& -- & --
& -- & --
& -- & --
& -- & --
& -- & -- \\

& ViSTA-SLAM
& \pmv{59.21}{0.02} & \pmv{14.56}{0.01}
& \pmv{81.71}{{<}0.01} & \pmv{10.71}{{<}0.01}
& \pmv{68.71}{{<}0.01} & \pmv{12.80}{{<}0.01}
& \pmv{24.39}{{<}0.01} & \pmv{8.48}{0.01}
& \pmv{40.38}{0.01} & \pmv{5.62}{0.01}
& \pmv{44.90}{0.02} & \pmv{12.89}{0.93}
& 53.22 & 10.84 \\

& VGGT-SLAM(Sim3)
& \pmv{18.16}{{<}0.01} & \pmv{3.24}{{<}0.01}
& \pmv{60.83}{{<}0.01} & \pmv{2.74}{{<}0.01}
& \pmv{26.01}{{<}0.01} & \pmv{3.41}{{<}0.01}
& \pmv{9.33}{{<}0.01} & \pmv{1.49}{{<}0.01}
& \pmv{11.49}{{<}0.01} & \pmv{0.44}{{<}0.01}
& \cellcolor{best}\pmv{3.87}{{<}0.01} & \pmv{1.06}{{<}0.01}
& 21.61 & 2.06 \\

& VGGT-SLAM2.0
& \pmv{32.99}{{<}0.01} & \pmv{4.98}{{<}0.01}
& \pmv{61.82}{{<}0.01} & \pmv{2.70}{{<}0.01}
& \pmv{39.10}{{<}0.01} & \pmv{4.02}{{<}0.01}
& \pmv{13.31}{{<}0.01} & \pmv{1.88}{{<}0.01}
& \pmv{16.67}{{<}0.01} & \pmv{0.41}{{<}0.01}
& \pmv{8.90}{{<}0.01} & \pmv{1.98}{{<}0.01}
& 28.80 & 2.66 \\

& VGGT-Long
& \cellcolor{second}\pmv{4.31}{{<}0.01} & \pmv{1.09}{{<}0.01}
& \cellcolor{best}\pmv{13.58}{{<}0.01} & \pmv{1.65}{{<}0.01}
& \cellcolor{second}\pmv{15.70}{{<}0.01} & \pmv{1.73}{{<}0.01}
& \pmv{7.92}{{<}0.01} & \pmv{1.60}{{<}0.01}
& \pmv{4.99}{{<}0.01} & \pmv{0.54}{{<}0.01}
& \pmv{5.28}{{<}0.01} & \pmv{1.08}{{<}0.01}
& \cellcolor{best}8.63 & 1.28 \\

\midrule
\multirow{10}{*}{\makecell[c]{w/o LC\\(Odom.)}}
& DSO
& \pmv{33.47}{0.20} & \pmv{0.27}{0.01}
& \pmv{19.49}{25.95} & \pmv{0.66}{1.02}
& \pmv{26.41}{1.04} & \pmv{0.45}{0.54}
& \pmv{1.74}{0.09} & \cellcolor{best}\pmv{0.17}{{<}0.01}
& -- & --
& \pmv{16.27}{0.38} & \cellcolor{best}\pmv{0.18}{{<}0.01}
& 19.48 & 0.35 \\

& ORB-SLAM3
& \pmv{22.89}{0.43} & \cellcolor{best}\pmv{0.26}{{<}0.01}
& \pmv{76.45}{15.73} & \pmv{1.15}{0.50}
& \cellcolor{best}\pmv{6.34}{2.06} & \cellcolor{best}\pmv{0.22}{{<}0.01}
& \cellcolor{best}\pmv{0.96}{0.14} & \pmv{0.17}{0.01}
& \cellcolor{second}\pmv{0.88}{0.14} & \cellcolor{second}\pmv{0.21}{0.04}
& \pmv{12.70}{0.40} & \pmv{0.21}{0.01}
& 20.04 & 0.37 \\

& DROID-SLAM
& \pmv{26.52}{0.02} & \pmv{0.30}{{<}0.01}
& \pmv{17.83}{0.76} & \cellcolor{best}\pmv{0.11}{{<}0.01}
& \pmv{32.37}{1.04} & \pmv{0.25}{0.01}
& \cellcolor{second}\pmv{1.37}{{<}0.01} & \pmv{0.24}{{<}0.01}
& \pmv{1.54}{{<}0.01} & \pmv{0.36}{{<}0.01}
& \pmv{22.65}{{<}0.01} & \pmv{0.27}{{<}0.01}
& 17.05 & \cellcolor{second}0.26 \\

\addlinespace[0.25em]
& MASt3R-SLAM
& -- & --
& -- & --
& -- & --
& -- & --
& -- & --
& -- & --
& -- & -- \\

& ViSTA-SLAM
& \pmv{59.59}{0.03} & \pmv{14.52}{{<}0.01}
& \pmv{81.71}{{<}0.01} & \pmv{10.71}{{<}0.01}
& \pmv{68.71}{{<}0.01} & \pmv{12.80}{{<}0.01}
& \pmv{24.39}{{<}0.01} & \pmv{8.48}{0.02}
& \pmv{28.20}{{<}0.01} & \pmv{6.93}{0.02}
& \pmv{48.82}{0.05} & \pmv{9.86}{0.24}
& 51.90 & 10.55 \\

& VGGT-SLAM(Sim3)
& \cellcolor{second}\pmv{19.84}{{<}0.01} & \pmv{3.17}{{<}0.01}
& \pmv{60.83}{{<}0.01} & \pmv{2.74}{{<}0.01}
& \pmv{26.81}{{<}0.01} & \pmv{3.17}{{<}0.01}
& \pmv{9.33}{{<}0.01} & \pmv{1.49}{{<}0.01}
& \pmv{11.49}{{<}0.01} & \pmv{0.43}{{<}0.01}
& \cellcolor{second}\pmv{7.00}{{<}0.01} & \pmv{1.47}{{<}0.01}
& 22.55 & 2.08 \\

& VGGT-SLAM2.0
& \pmv{20.10}{{<}0.01} & \pmv{3.55}{{<}0.01}
& \pmv{61.82}{{<}0.01} & \pmv{2.70}{{<}0.01}
& \pmv{31.64}{{<}0.01} & \pmv{3.80}{{<}0.01}
& \pmv{13.31}{{<}0.01} & \pmv{1.88}{{<}0.01}
& \pmv{16.67}{{<}0.01} & \pmv{0.41}{{<}0.01}
& \pmv{26.33}{{<}0.01} & \pmv{4.52}{{<}0.01}
& 28.31 & 2.81 \\

& VGGT-Long
& \cellcolor{best}\pmv{17.04}{{<}0.01} & \pmv{1.50}{{<}0.01}
& \cellcolor{second}\pmv{13.58}{{<}0.01} & \pmv{1.65}{{<}0.01}
& \cellcolor{second}\pmv{21.94}{{<}0.01} & \pmv{2.23}{{<}0.01}
& \pmv{7.92}{{<}0.01} & \pmv{1.60}{{<}0.01}
& \pmv{4.99}{{<}0.01} & \pmv{0.54}{{<}0.01}
& \cellcolor{best}\pmv{6.89}{{<}0.01} & \pmv{1.75}{{<}0.01}
& \cellcolor{best}12.06 & 1.54 \\

\addlinespace[0.25em]
& VO+DA3
& -- & --
& -- & --
& \pmv{33.59}{1.44} & \pmv{0.24}{0.05}
& \pmv{2.28}{0.34} & \cellcolor{second}\pmv{0.17}{0.01}
& \pmv{0.98}{0.05} & \pmv{0.23}{0.03}
& \pmv{30.53}{1.37} & \pmv{0.20}{0.02}
& -- & -- \\

& \textbf{Ours}
& \pmv{33.24}{0.38} & \cellcolor{second}\pmv{0.27}{0.01}
& \cellcolor{best}\pmv{3.65}{1.05} & \cellcolor{second}\pmv{0.13}{0.03}
& \pmv{30.18}{0.98} & \cellcolor{second}\pmv{0.23}{0.01}
& \pmv{1.86}{0.30} & \pmv{0.18}{0.01}
& \cellcolor{best}\pmv{0.62}{0.09} & \cellcolor{best}\pmv{0.20}{0.02}
& \pmv{23.03}{0.98} & \cellcolor{second}\pmv{0.20}{{<}0.01}
& \cellcolor{second}15.43 & \cellcolor{best}0.20 \\
\bottomrule
\end{tabular}%
}

\endgroup
\end{table*}

%% file: tables/Table_KITTI_06-10.tex
\begin{table*}[!htbp]
\centering
\caption{KITTI odometry results on Seq. 06--10 using $t_{\rm rel}$ (\%) and $r_{\rm rel}$ (deg/100m).}
\vspace{-0.3cm}
\label{tab:kitti_06_10}
\small
\begingroup
\setlength{\tabcolsep}{1.5pt}
\renewcommand{\arraystretch}{1.05}

\resizebox{\textwidth}{!}{%
\begin{tabular}{@{}llcccccccccccc@{}}
\toprule
\multirow{2}{*}{Mode} & \multirow{2}{*}{Method}
& \multicolumn{2}{c}{\textbf{06}}
& \multicolumn{2}{c}{\textbf{07}}
& \multicolumn{2}{c}{\textbf{08}}
& \multicolumn{2}{c}{\textbf{09}}
& \multicolumn{2}{c}{\textbf{10}}
& \multicolumn{2}{c}{\textbf{Avg.}} \\
\cmidrule(lr){3-4}\cmidrule(lr){5-6}\cmidrule(lr){7-8}\cmidrule(lr){9-10}\cmidrule(lr){11-12}\cmidrule(lr){13-14}
& & $t_{\rm rel}$ & $r_{\rm rel}$ & $t_{\rm rel}$ & $r_{\rm rel}$ & $t_{\rm rel}$ & $r_{\rm rel}$ & $t_{\rm rel}$ & $r_{\rm rel}$ & $t_{\rm rel}$ & $r_{\rm rel}$ & $t_{\rm rel}$ & $r_{\rm rel}$ \\
\midrule
\multirow{7}{*}{w/ LC} & ORB-SLAM3 & \pmv{6.78}{0.31} & \cellcolor{best}\pmv{0.19}{0.01} & \cellcolor{best}\pmv{3.21}{1.02} & \cellcolor{second}\pmv{0.43}{0.12} & \cellcolor{second}\pmv{16.20}{1.04} & \cellcolor{best}\pmv{0.28}{{<}0.01} & \cellcolor{best}\pmv{3.98}{0.70} & \cellcolor{best}\pmv{0.35}{0.06} & \cellcolor{best}\pmv{3.06}{0.32} & \cellcolor{best}\pmv{0.25}{0.02} & \cellcolor{second}6.65 & \cellcolor{second}0.30 \\
 & DROID-SLAM & \pmv{20.20}{0.16} & \cellcolor{second}\pmv{0.24}{0.02} & \pmv{15.30}{0.27} & \cellcolor{best}\pmv{0.32}{{<}0.01} & \pmv{26.10}{0.21} & \cellcolor{second}\pmv{0.32}{{<}0.01} & \pmv{22.36}{0.42} & \cellcolor{second}\pmv{0.39}{0.14} & \cellcolor{second}\pmv{6.75}{0.07} & \cellcolor{second}\pmv{0.28}{{<}0.01} & 18.14 & \cellcolor{best}0.31 \\
\addlinespace[0.25em]
 & MASt3R-SLAM & -- & -- & -- & -- & -- & -- & -- & -- & -- & -- & -- & -- \\
 & ViSTA-SLAM & \pmv{43.36}{1.29} & \pmv{6.02}{0.10} & \pmv{17.72}{{<}0.01} & \pmv{9.58}{{<}0.01} & \pmv{62.00}{{<}0.01} & \pmv{10.98}{{<}0.01} & \pmv{78.34}{{<}0.01} & \pmv{5.99}{{<}0.01} & \pmv{54.95}{{<}0.01} & \pmv{9.26}{{<}0.01} & 51.27 & 8.37 \\
 & VGGT-SLAM(Sim3) & \pmv{7.75}{{<}0.01} & \pmv{1.31}{{<}0.01} & \pmv{40.31}{{<}0.01} & \pmv{7.82}{{<}0.01} & \pmv{20.69}{{<}0.01} & \pmv{1.63}{{<}0.01} & \pmv{36.38}{{<}0.01} & \pmv{8.73}{{<}0.01} & \pmv{16.32}{{<}0.01} & \pmv{4.56}{{<}0.01} & 24.29 & 4.81 \\
 & VGGT-SLAM2.0 & \cellcolor{second}\pmv{4.73}{{<}0.01} & \pmv{2.27}{{<}0.01} & \pmv{37.92}{{<}0.01} & \pmv{3.76}{{<}0.01} & \pmv{16.90}{{<}0.01} & \pmv{2.52}{{<}0.01} & \pmv{22.55}{{<}0.01} & \pmv{3.00}{{<}0.01} & \pmv{17.18}{{<}0.01} & \pmv{5.96}{{<}0.01} & 19.86 & 3.50 \\
 & VGGT-Long & \cellcolor{best}\pmv{3.14}{{<}0.01} & \pmv{0.59}{{<}0.01} & \cellcolor{second}\pmv{4.00}{{<}0.01} & \pmv{1.62}{{<}0.01} & \cellcolor{best}\pmv{16.13}{{<}0.01} & \pmv{1.54}{{<}0.01} & \cellcolor{second}\pmv{12.26}{{<}0.01} & \pmv{1.95}{{<}0.01} & \pmv{17.81}{{<}0.01} & \pmv{2.84}{{<}0.01} & \cellcolor{best}10.67 & 1.71 \\

\midrule
\multirow{10}{*}{\makecell[c]{w/o LC\\(Odom.)}} & DSO & \pmv{21.52}{0.40} & \pmv{0.21}{{<}0.01} & \pmv{11.65}{0.67} & \pmv{0.37}{{<}0.01} & \pmv{31.56}{0.27} & \cellcolor{best}\pmv{0.25}{{<}0.01} & \cellcolor{second}\pmv{14.34}{0.09} & \cellcolor{best}\pmv{0.20}{{<}0.01} & \cellcolor{second}\pmv{6.22}{0.25} & \cellcolor{second}\pmv{0.26}{{<}0.01} & 17.06 & 0.26 \\
 & ORB-SLAM3 & \pmv{18.06}{0.70} & \pmv{0.24}{0.01} & \cellcolor{second}\pmv{10.94}{0.22} & \cellcolor{second}\pmv{0.35}{0.02} & \pmv{17.34}{0.61} & \cellcolor{second}\pmv{0.28}{{<}0.01} & \cellcolor{best}\pmv{13.53}{0.23} & \pmv{0.24}{0.01} & \cellcolor{best}\pmv{3.12}{0.27} & \pmv{0.27}{0.01} & \cellcolor{second}12.60 & 0.28 \\
 & DROID-SLAM & \pmv{19.90}{0.02} & \pmv{0.24}{{<}0.01} & \pmv{15.58}{0.33} & \cellcolor{best}\pmv{0.31}{{<}0.01} & \pmv{26.37}{0.21} & \pmv{0.32}{{<}0.01} & \pmv{21.54}{0.41} & \pmv{0.47}{0.26} & \pmv{6.73}{0.11} & \pmv{0.27}{{<}0.01} & 18.02 & 0.32 \\
\addlinespace[0.25em]
 & MASt3R-SLAM & -- & -- & -- & -- & -- & -- & -- & -- & -- & -- & -- & -- \\
 & ViSTA-SLAM & \pmv{42.92}{2.08} & \pmv{5.43}{0.58} & \pmv{17.98}{0.23} & \pmv{8.77}{0.69} & \pmv{62.00}{{<}0.01} & \pmv{10.98}{{<}0.01} & \pmv{78.60}{0.23} & \pmv{5.72}{0.23} & \pmv{54.95}{{<}0.01} & \pmv{9.26}{{<}0.01} & 51.29 & 8.03 \\
 & VGGT-SLAM(Sim3) & \cellcolor{second}\pmv{13.15}{{<}0.01} & \pmv{1.78}{{<}0.01} & \pmv{61.46}{{<}0.01} & \pmv{28.18}{{<}0.01} & \pmv{20.69}{{<}0.01} & \pmv{1.63}{{<}0.01} & \pmv{31.10}{{<}0.01} & \pmv{1.73}{{<}0.01} & \pmv{16.32}{{<}0.01} & \pmv{4.56}{{<}0.01} & 28.54 & 7.58 \\
 & VGGT-SLAM2.0 & \pmv{15.50}{{<}0.01} & \pmv{1.73}{{<}0.01} & \pmv{39.20}{{<}0.01} & \pmv{2.58}{{<}0.01} & \cellcolor{second}\pmv{16.90}{{<}0.01} & \pmv{2.52}{{<}0.01} & \pmv{28.28}{{<}0.01} & \pmv{1.99}{{<}0.01} & \pmv{17.18}{{<}0.01} & \pmv{5.96}{{<}0.01} & 23.41 & 2.96 \\
 & VGGT-Long & \cellcolor{best}\pmv{6.57}{{<}0.01} & \pmv{1.00}{{<}0.01} & \cellcolor{best}\pmv{9.21}{{<}0.01} & \pmv{1.90}{{<}0.01} & \cellcolor{best}\pmv{16.13}{{<}0.01} & \pmv{1.54}{{<}0.01} & \pmv{18.63}{{<}0.01} & \pmv{2.19}{{<}0.01} & \pmv{17.81}{{<}0.01} & \pmv{2.84}{{<}0.01} & \cellcolor{best}13.67 & 1.89 \\
\addlinespace[0.25em]
 & VO+DA3 & \pmv{31.16}{5.13} & \cellcolor{second}\pmv{0.19}{0.02} & \pmv{13.54}{0.65} & \pmv{0.37}{0.10} & \pmv{38.01}{2.87} & \pmv{0.37}{0.10} & \pmv{25.79}{0.66} & \pmv{0.28}{0.01} & \pmv{8.19}{0.77} & \cellcolor{best}\pmv{0.24}{0.01} & 23.34 & \cellcolor{second}0.29 \\
 & \textbf{Ours} & \pmv{23.96}{0.96} & \cellcolor{best}\pmv{0.19}{0.02} & \pmv{11.54}{0.42} & \pmv{0.35}{0.02} & \pmv{36.44}{0.91} & \pmv{0.35}{0.02} & \pmv{21.42}{0.11} & \cellcolor{second}\pmv{0.23}{{<}0.01} & \pmv{7.26}{0.32} & \pmv{0.32}{{<}0.01} & 20.12 & \cellcolor{best}0.29 \\
\bottomrule
\end{tabular}%
}
\endgroup
\vspace{-0.3cm}
\end{table*}

%% file: tables/Table_speed_compare.tex
\definecolor{best}{HTML}{A7D0A7}
\definecolor{second}{HTML}{E2EFE2}

\begin{table}[t]
\centering
\caption{Runtime and GPU memory on EuRoC MH sequences.}
\vspace{-0.2cm}
\label{table:system_speed}
\footnotesize
\setlength{\tabcolsep}{1.4pt}
\renewcommand{\arraystretch}{1.08}
\begin{tabular*}{0.48\textwidth}{@{\extracolsep{\fill}}lcccccc@{}}
\toprule
\multirow{2}{*}{\textbf{Method}} &
\multicolumn{3}{c}{\textbf{RTX 4090}} &
\multicolumn{3}{c}{\textbf{RTX 4060 Laptop}} \\
\cmidrule(lr){2-4}\cmidrule(lr){5-7}
&
\makecell{\textbf{Pose}\\\textbf{(Hz)}} &
\makecell{\textbf{Map}\\\textbf{(Hz)}} &
\makecell{\textbf{VRAM}\\\textbf{(GB)}} &
\makecell{\textbf{Pose}\\\textbf{(Hz)}} &
\makecell{\textbf{Map}\\\textbf{(Hz)}} &
\makecell{\textbf{VRAM}\\\textbf{(GB)}} \\
\midrule
DROID-SLAM     & \cellcolor{best}52.6 & 19.0 & 11.4 & OOM & OOM & OOM \\ 
MASt3R-SLAM    & 24.4 & \cellcolor{best}24.4 & 33.4 & OOM & OOM & OOM \\
ViSTA-SLAM     & 18.7 & 18.7 & 10.2 & OOM & OOM & OOM \\
VGGT-SLAM      & 21.7 & 21.7 & 11.9 & OOM & OOM & OOM \\
VGGT-SLAM 2.0  & 22.2 & \cellcolor{second}22.2 & \cellcolor{second}9.0 & \cellcolor{second}3.3 & \cellcolor{best}3.3 & \cellcolor{second}6.6 \\
Ours           & \cellcolor{second}34.2 & 21.0 & \cellcolor{best}8.5 & \cellcolor{best}16.1 & \cellcolor{second}3.2 & \cellcolor{best}6.5 \\
\bottomrule
\end{tabular*}
\begin{flushleft}
\scriptsize $^*$ OOM denotes out-of-memory failure.
\end{flushleft}
\vspace{-0.3cm}
\end{table}

%% file: tables/Table_recover.tex
\begin{table}[!t]
\centering
\caption{Pose-output rate and VGGT invocations on EuRoC MH sequences. Invocations are reported over sequence length.}
\label{tab:recovery_fps}
\vspace{-0.25cm}
\scriptsize
\setlength{\tabcolsep}{3.2pt}
\renewcommand{\arraystretch}{1.05}

\begin{tabular*}{\columnwidth}{@{\extracolsep{\fill}}lccccc@{}}
\toprule
\textbf{Metric} & \textbf{MH01} & \textbf{MH02} & \textbf{MH03} & \textbf{MH04} & \textbf{MH05} \\
\midrule
Pose (Hz) & 19.1 & 19.0 & 18.7 & 10.4 & 13.3 \\
VGGT invocations / frames & $2/3682$ & $2/3040$ & $1/2700$ & $30/2033$ & $15/2273$ \\
\bottomrule
\end{tabular*}
\vspace{-0.4cm}
\end{table}

%% file: tables/Table_runtime_analysis.tex
\begin{table}[htbp]
  \centering
  \caption{Average Execution Time of Individual Modules}
  \vspace{-0.3cm}
  \label{tab:timing_analysis}
  \begin{threeparttable}
  \renewcommand{\arraystretch}{1.1}
  \begin{tabular*}{0.48\textwidth}{@{  } @{\extracolsep{\fill}} l l c @{  }}
    \toprule
    \textbf{Thread} & \textbf{Module} & \textbf{Time (ms)} \\
    \midrule
    
    \multirow{3}{*}{Frontend} 
        & Tracking & $54.11$ \\
        & Keyframe Creation & $2.33$ \\
        & \textit{Total Latency} & $56.44$ \\
    \midrule
    
    \multirow{2}{*}{Backend} 
        & First Stage Opt. & $32.70$ \\
        & Second Stage Opt. & $9.98$ \\
    \midrule
    
    \multirow{2}{*}{Async. VGGT} 
        & VGGT Inference\tnote{*} & $2807.56$ \\
        & Pose Graph Opt. & $5.14$ \\
    
    \bottomrule
  \end{tabular*}
  
  \begin{tablenotes}
    \footnotesize
    \item[*] Denotes the execution time for a single inference pass, which processes a batch of 10 keyframes.
  \end{tablenotes}
  \end{threeparttable}
  \vspace{-0.3cm}
\end{table}

%% file: article/conclusion.tex
In this paper, we propose HyVGGT-VO, a hybrid framework that combines the efficiency of sparse VO with the dense mapping capability of feed-forward models. Under the odometry-only setting, HyVGGT-VO outperforms VGGT-SLAM 2.0 on EuRoC MAV, TUM RGB-D, and KITTI, reducing the average ATE by 91\% and 20\% on EuRoC MAV and TUM RGB-D, respectively, and decreasing the KITTI Seq. 00–05 relative translation errors and relative rotation errors by 46\% and 92\%. Overall, the system provides continuous high-rate pose estimates and dense mapping within a constrained GPU memory budget. Future work will explore dynamic-aware hybrid tracking using feed-forward cues for KLT filtering and backend weighting.